\documentclass[conference]{IEEEtran}
\IEEEoverridecommandlockouts
\usepackage{cite}
\usepackage{amsmath,amssymb,amsfonts}
\usepackage{algorithmic}
\usepackage{graphicx}
\usepackage{textcomp}
\usepackage{xcolor}
\usepackage{threeparttable}

\usepackage{multirow}
\usepackage{subfigure}
\usepackage{algorithm}
\usepackage{xspace}
\usepackage{dsfont}
\usepackage{epsfig}
\usepackage{epstopdf}
\usepackage{booktabs}
\usepackage{indentfirst}
\usepackage{amsmath}
\usepackage{amsthm}
\usepackage{bm}
\usepackage{listings}
\usepackage{tabu}
\usepackage{longtable}
\usepackage{threeparttable}
\usepackage{marvosym}
\usepackage{bigstrut}
\usepackage{hyperref}
\usepackage{soul}

\usepackage{tabularx}
\newtheorem{mydef}{Definition}

\newcommand{\name}{START\xspace}
\newcommand{\gat}{TPE-GAT\xspace}
\newcommand{\enc}{TAT-Enc\xspace}
\newcommand{\bj}{BJ\xspace}
\newcommand{\porto}{Porto\xspace}
\newcommand{\geolife}{Geolife\xspace}

\newcommand{\ie}{\emph{i.e.,}\xspace}

\newcommand{\eg}{\emph{e.g.,}\xspace}
\newcommand{\etal}{\emph{et al.}\xspace}
\newcommand{\ignore}[1]{}

\newcommand{\changed}[1]{\textcolor{black}{#1}}

\def\BibTeX{{\rm B\kern-.05em{\sc i\kern-.025em b}\kern-.08em
    T\kern-.1667em\lower.7ex\hbox{E}\kern-.125emX}}
\begin{document}

\title{Self-supervised Trajectory Representation Learning with Temporal Regularities and Travel Semantics}


\author{
    \IEEEauthorblockN{Jiawei Jiang\textsuperscript{1,2}, Dayan Pan\textsuperscript{1,2}, Houxing Ren\textsuperscript{1,2}, Xiaohan Jiang\textsuperscript{1}, Chao Li\textsuperscript{1,2}, Jingyuan Wang\textsuperscript{1,3,4,${\ast}$}\thanks{$^{\ast}$ Corresponding Author: Jingyuan Wang}}
    \IEEEauthorblockA{$^1$School of Computer Science and Engineering, Beihang University, Beijing, China}
    \IEEEauthorblockA{$^2$Z-park Strategic Alliance of Smart City Industrial Technology Innovation, Beijing, China}
    \IEEEauthorblockA{$^3$Pengcheng Laboratory, Shenzhen, China}
    \IEEEauthorblockA{$^4$School of Economics and Management, Beihang University, Beijing, China}
    \IEEEauthorblockA{\{jwjiang, dayan, renhouxing, jxh199, licc, jywang\}@buaa.edu.cn}
}

\maketitle

\begin{abstract}
\changed{Trajectory Representation Learning (TRL) is a powerful tool for spatial-temporal data analysis and management. TRL aims to convert complicated raw trajectories into low-dimensional representation vectors, which can be applied to various downstream tasks, such as trajectory classification, clustering, and similarity computation. Existing TRL works usually treat trajectories as ordinary sequence data, while some important spatial-temporal characteristics, such as temporal regularities and travel semantics, are not fully exploited. To fill this gap, we propose a novel \underline{S}elf-supervised trajectory representation learning framework with \underline{T}empor\underline{A}l \underline{R}egularities and \underline{T}ravel semantics, namely \name. The proposed method consists of two stages. The first stage is a Trajectory Pattern-Enhanced Graph Attention Network (\gat), which converts the road network features and travel semantics into representation vectors of road segments. The second stage is a Time-Aware Trajectory Encoder (\enc), which encodes representation vectors of road segments in the same trajectory as a trajectory representation vector, meanwhile incorporating temporal regularities with the trajectory representation. Moreover, we also design two self-supervised tasks, \ie span-masked trajectory recovery and trajectory contrastive learning, to introduce spatial-temporal characteristics of trajectories into the training process of our \name framework. The effectiveness of the proposed method is verified by extensive experiments on two large-scale real-world datasets for three downstream tasks. The experiments also demonstrate that our method can be transferred across different cities to adapt heterogeneous trajectory datasets.}
\end{abstract}



\section{Introduction}

With the rapid development of GPS-enabled devices, a large amount of trajectory data can be collected in cities. Trajectory data analysis and management, such as trajectory-based prediction~\cite{mdtp, astar}, traffic prediction~\cite{libcity, stden}, urban dangerous goods management~\cite{good1}, and trajectory similarity computation~\cite{similarity}, have become a hot topic in the data engineering community. Traditional research on trajectory data analysis requires manual feature engineering and unique models for specific tasks, making them difficult to transfer to different applications~\cite{fu2020trembr}. To improve the generality of tools for analyzing trajectory data analysis tools, \emph{Trajectory Representation Learning (TRL)} has emerged in recent years~\cite{li2018deep, yao2017trajectory}. TRL aims to transform raw trajectories into generic low-dimensional representation vectors that can be applied in various downstream tasks rather than being limited to a specific task.


\changed{In the literature, earlier TRL studies directly use general sequence-to-sequence models (such as LSTM~\cite{lstm} and Transformers~\cite{vaswani2017attention}) with reconstruction tasks to generate trajectory representation vectors~\cite{li2018deep, yao2017trajectory, fu2020trembr}. Such models consider trajectories as ordinary sequence data and thus cannot fully capture spatial-temporal semantic information of trajectories in the representation vectors. After this, many trajectory representation learning methods are proposed for specific downstream tasks, such as for approximate trajectory similarity computation~\cite{yao2019computing,yang2021t3s}, trajectory clustering~\cite{fang20212}, anomalous trajectory detection~\cite{abnormal} and path ranking~\cite{prank}.}

\changed{In recent years, some two-stage methods have been proposed to learn generic trajectory representations for multiple downstream tasks~\cite{chen2021robust,yang2021unsupervised}. These methods first adopt a graph representation learning to convert road segments of a static road network into representation vectors and then use sequential deep learning models with self-supervised tasks to convert the road representation vectors in the same trajectory into a trajectory representation vector. For example, Toast~\cite{chen2021robust} and PIM~\cite{yang2021unsupervised} use node2vec~\cite{grover2016node2vec} to learn road representations and respectively use Transformer with masked prediction and RNN with mutual information maximization as self-supervised tasks to generate generic trajectory representations. These two-stage methods incorporate the static road network as spatial semantic information in the trajectory representations so they can improve downstream tasks. However, trajectory data contains rather complicated spatial-temporal semantic information. Many critical spatial-temporal characteristics and semantic information are helpful for downstream tasks but are still not fully utilized by existing works.}

\begin{figure}[t]
    \centering
    \subfigure[Trajectory Frequencies.]{
        \includegraphics[width=0.29\columnwidth, page=1]{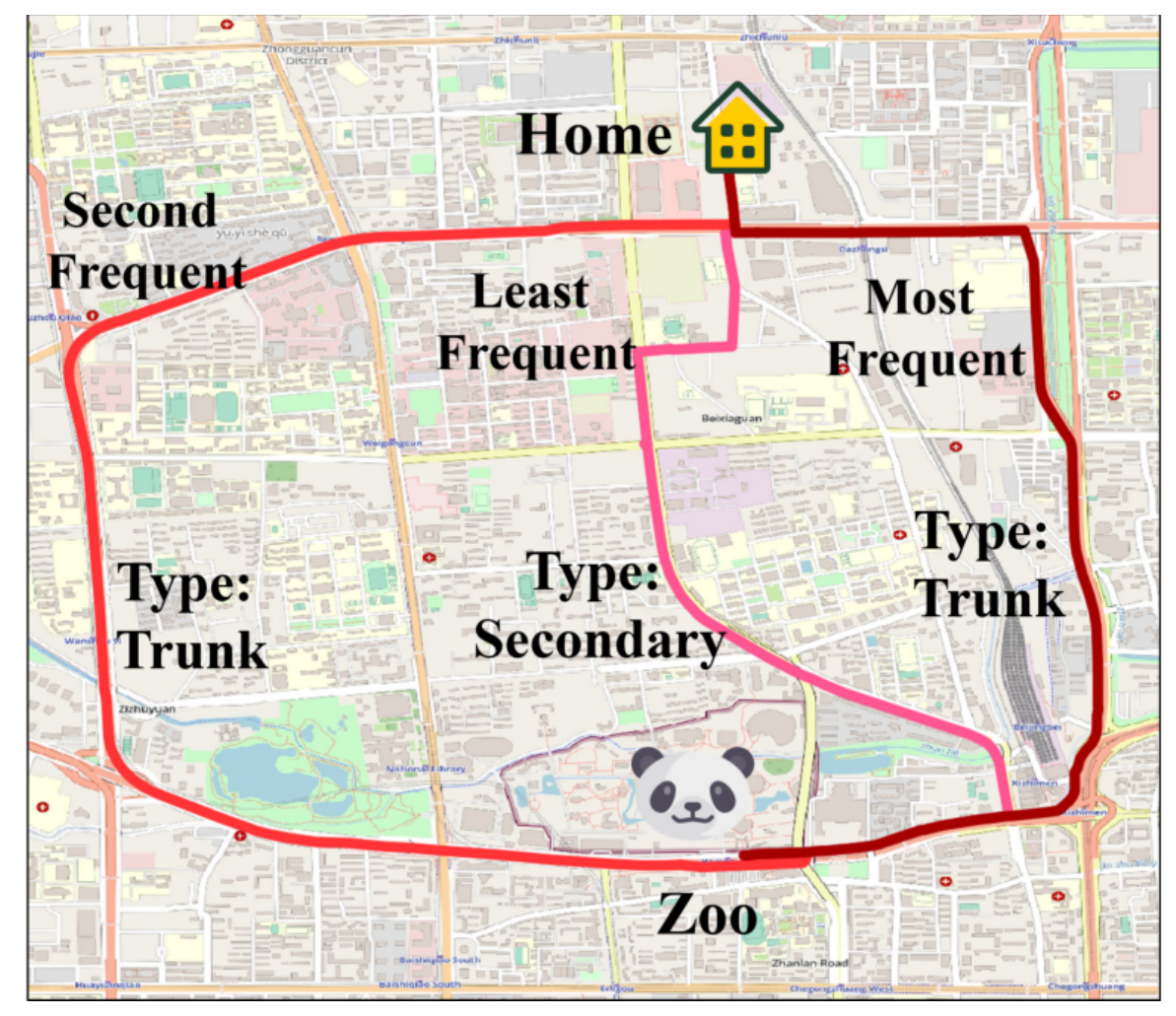}
        \label{fig:introa}
    }
    \subfigure[Periodic Patterns of Urban Traffic.]{
        \includegraphics[width=0.3\columnwidth, page=2, page=1]{figure/traj0.pdf}
        \label{fig:introb}
    }
    \subfigure[Time Interval Distribution.]{
        \includegraphics[width=0.29\columnwidth, page=3, page=1]{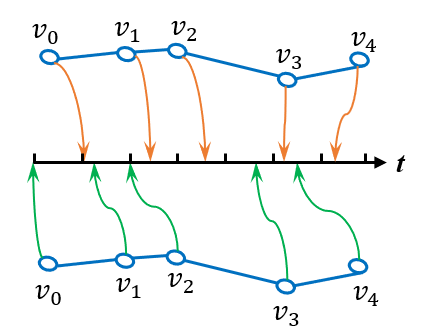}
        \label{fig:introc}
    }
    \vspace{-.2cm}
    \caption{\changed{Temporal Regularities and Travel Semantics in Trajectories. (Map data $\copyright$ OpenStreetMap contributors, CC BY-SA.)}}
    \label{fig:intro}
    \vspace{-.7cm}
\end{figure}

\changed{The first characteristic that should be considered in TRL is travel semantics. As shown in Figure~\ref{fig:introa}, road segments traversed by different trajectories with the same origin and destination (OD) have different road types and visit frequencies, \ie the human mobility patterns. Both of these travel-related semantics are useful for downstream tasks and should be incorporated into trajectory representations. However, previous works such as Toast~\cite{chen2021robust} and PIM~\cite{yang2021unsupervised} only model the static road network information in their representation learning but fail to incorporate the travel semantic information. The second characteristic that should be considered is temporal regularities. From a macro perspective, the trajectories generated by vehicles in the city are influenced by the periodic temporal patterns of urban traffic. As shown in Figure~\ref{fig:introb}, the number of urban trajectories exhibits an apparent periodic pattern, \ie the number of trajectories during the morning and evening rush hours is much larger than usual. A large number of trajectories means congested road conditions, which naturally affects the generation of trajectories. From a micro perspective, irregular time intervals are another temporal regularity of trajectories. As shown in Figure~\ref{fig:introc}, for two trajectories with the same shape, the sample points, \ie road segments, can be distributed quite differently on the time axis. It is because the travel time of a road segment is dynamic, which can also reflect the congestion level of roads. Both temporal regularities of periodic patterns and irregular time intervals are useful for downstream tasks. However, most previous works consider trajectories only as sequences of locations~\cite{li2018deep,chen2021robust,yang2021unsupervised} and do not consider the temporal information in their methods. In addition, the self-supervised tasks in existing TRL methods do not sufficiently consider the spatial-temporal characteristics of the trajectories. Most methods use general sequence reconstruction~\cite{li2018deep,yao2017trajectory} or masked prediction tasks (MLM)~\cite{chen2021robust} as their self-supervised tasks, which treat trajectories as general sequence data and fail to capture temporal and travel semantics. This problem limits the performance of the learned representation vectors on downstream tasks.}

\changed{In this paper, we propose a novel \textbf{S}elf-supervised trajectory representation learning framework with \textbf{T}empor\textbf{A}l \textbf{R}egularities and \textbf{T}ravel semantics, abbreviated as \name. The framework integrates temporal regularities and travel semantics into TRL using a two-stage learning method. The first stage is a Trajectory Pattern-Enhanced Graph Attention Network (\gat), which converts a road network into road segment representation vectors. Travel semantics information is also incorporated at this stage. Specifically, the \gat module takes rich road features as input and extends Graph Attention Network~\cite{velivckovic2017graph} using a road segment transfer probability matrix to model road visit frequencies. In this way, both the travel semantics of road features and visit frequencies are integrated into the road representations. The second stage converts road representation sequences into trajectory representations and incorporates temporal regularity information. Here we propose a Time-Aware Trajectory Encoder Layer (\enc) to incorporate temporal regularities. Specifically, the \enc fuses minute and day-of-week indexes with road segment representations to capture the periodic temporal patterns of urban traffic and adopts a Time Interval-Aware Self-Attention to process irregular time interval information.}

\changed{We also design two self-supervised tasks to train our \name. The first is \emph{span-masked trajectory recovery}, which masks consecutive subsequences in trajectories to capture the local features and order information. The second is \emph{trajectory contrastive learning}, which employs four data augmentation methods that consider the spatial-temporal characteristics of trajectories to train the contrastive learning loss. Compared to the traditional self-supervised tasks such as sequence reconstruction and MLM, the proposed tasks fully consider the spatial-temporal characteristics of the trajectories. The effectiveness of the proposed method is verified by extensive experiments on two large-scale datasets for three downstream tasks. The results show that \name significantly outperforms the state-of-the-art models.}

\changed{In summary, the main contributions of this paper are summarized as follows:}
\changed{
\begin{itemize}
    \item We propose a two-stage TRL method incorporating temporal regularities and travel semantics into trajectory representations. Compared to previous TRL research, the proposed method can utilize more spatial-temporal characteristics of trajectories for downstream tasks.
    \item We design two self-supervised tasks to train our \name. Compared to traditional self-supervised tasks for general sequence representation learning, such as sequence reconstruction and MLM, the proposed tasks are more suitable for TRL since they account for the spatial-temporal characteristics of trajectories. We believe these tasks can be applied to other TRL model training.
    \item In addition to superior performance, the experiments also demonstrate that the proposed self-supervised tasks can use fewer data to outperform the supervised model. Moreover, our methods can be transferred across heterogeneous road network datasets. To the best of our knowledge, this is the first TRL method with this feature, which is very useful for solving the problem of insufficient data in many real-world applications.
\end{itemize}
}

\section{PRELIMINARIES}

In this section, we first introduce basic notations and preliminaries used in this paper. Then we formalize the problem of trajectory representation learning.

\subsection{Notations and Definitions}\label{nota:def}

\begin{mydef}[Road Network]
We represent road network as a directed graph $\mathcal{G} = (\mathcal{V}, \mathcal{E}, \bm{F}_\mathcal{V}, \bm{A})$, where $\mathcal{V} = \{v_1, \cdots, v_{|\mathcal{V}|}\}$ is a set of $|\mathcal{V}|$ vertices, each vertex $v_i$ representing a road segment, $\mathcal{N}_i$ is the neighborhood of road segment $v_i$, $\mathcal{E} \subseteq \mathcal{V} \times \mathcal{V}$ is a set of edges, each ${e_{i,j}} = (v_i, v_j)$ representing the intersection between road segments $v_i$ and $v_j$, $\bm{F}_\mathcal{V} \in \mathbb{R}^{|\mathcal{V}| \times d_{in}}$ is the features of road segments, and $\bm{A} \in \mathbb{R}^{|\mathcal{V}| \times |\mathcal{V}|}$ is a binary value adjacency matrix of network $\mathcal{G}$ indicating whether there exists a directed link between roads.
\end{mydef}

\begin{mydef}[GPS-based Trajectory]
A GPS-based trajectory (or a raw trajectory) $\mathcal{T}^{raw}$ is a sequence of spatial-temporal sample points recorded by GPS-enabled devices, a sample point $sp=\langle lat_i, lon_i, t_i \rangle$ is a triplet consisting of latitude, longitude, and a visit timestamp.
\end{mydef}

\begin{mydef}[Road-network Constrained Trajectory]
A road-network constrained trajectory $\mathcal{T}$ is a time-ordered sequence of $m$ adjacent road segments generated by a user, i.e., $\mathcal{T}=[\langle v_i, t_i \rangle]_{i=1}^m$, where $v_i \in \mathcal{V}$ presents the $i$-th road segment and $t_i$ is the visit timestamp for $v_i$. For simplicity, we also use \underline{roads} to refer to \underline{road segments} in the following.
\end{mydef}

In this study, we mainly focus on the road-network constrained trajectories. Therefore, given a raw trajectory $\mathcal{T}^{raw}$ and the road network $\mathcal{G}$, we perform the \textit{map matching}\cite{yang2018fast} procedure to align trajectory points with road segments and get the road-network constrained trajectory $\mathcal{T}$.

\subsection{Problem Statement}

Given a trajectory dataset $\mathcal{D}=\{{\mathcal{T}_i}\}_{i=1}^{|\mathcal{D}|}$ and a road network $\mathcal{G}$, the \textit{Trajectory Representation Learning} (TRL) task aims to learn a generic low-dimensional representation ${\bm{p}_i} \in \mathbb{R}^{d}$ for each trajectory $\mathcal{T}_{i} \in \mathcal{D}$. Specifically, in this study, we aim to develop a self-supervised framework that encodes each trajectory $\mathcal{T}_{i}$ into a generic $d$-dimensional representation vector ${\bm{p}_i}$, which can be applied in various downstream tasks, such as travel time estimation, trajectory classification, and trajectory similarity computation.


\section{Method}

\begin{figure}[t]
    \centering
    \includegraphics[width=0.95\columnwidth]{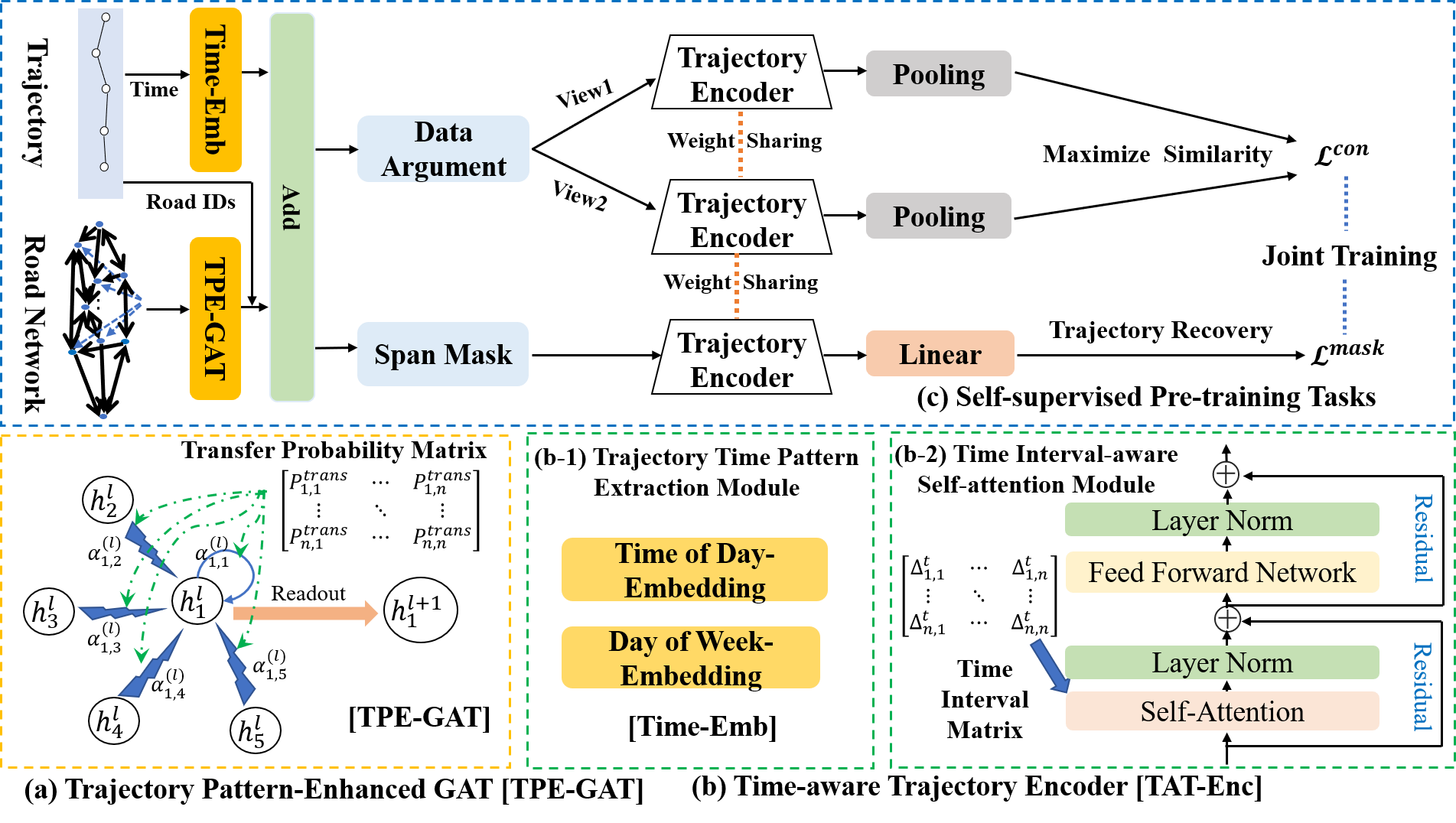}
    \vspace{-.3cm}
    \caption{Overall Framework of \name.}
    \label{fig:overall}
    \vspace{-.6cm}
\end{figure}

In this section, we introduce the proposed \name framework. Figure~\ref{fig:overall} provides an overview of it. We start with the framework structure, including a trajectory pattern-enhanced graph attention layer (\gat) and a time-aware trajectory encoder layer (\enc). Then, we present two self-supervised tasks to train \name. Finally, we display how to adapt the learned representations to specific downstream tasks.





\subsection{Trajectory Pattern-Enhanced Graph Attention Layer}\label{sec:gat}

The Trajectory Pattern-Enhanced Graph Attention Network (\gat) is the first stage of \name, which converts a road network into road representation vectors and incorporates the travel semantics of the trajectories. As mentioned in Section~\ref{nota:def}, the roads in the trajectory have some important inherent properties, and they are constrained by the connectivity of the road network. Therefore, we learn the road-level representation vector from both the road features and network structure. Previous works often use random walk-based models such as node2vec~\cite{grover2016node2vec} to encode the static road network as spatial semantic information used in the trajectory representations~\cite{li2018deep, chen2021robust}. However, such learning methods fail to incorporate road features and travel semantics in the trajectories, such as visit frequencies.

Therefore, we propose using graph neural networks to capture both the road features and network structure. Considering that the road network is a directed graph, we choose the graph attention network (GAT)~\cite{velivckovic2017graph} because it can dynamically assign weights to the neighborhood nodes by computing the attention weights between pairs of nodes. However, the standard GAT cannot capture the travel patterns in the trajectories. To solve this problem, we propose a Trajectory Pattern-Enhanced Graph Attention Network, namely \gat, which extends the computation of attention weights of GAT by introducing the transfer probability matrix between roads computed from the historical data to model visiting frequencies of roads.

The \gat consists of $L_1$ layers in total. First, we take rich road features $\bm{F}_\mathcal{V}$ as input to the first layer. Specifically, given a road $v_i$, we consider six types of features, namely road type, road length, number of lanes, maximum travel speed, in-degrees, and out-degrees in the road network. We concatenate these features to create the initialized road representation $\bm{h}_i^{(0)} \in \mathbb{R}^{d_0}$ for the road $v_i$. Then, the attention weight $\alpha_{ij}$ between road $v_i$ and $v_j$ in the $l$-th layer are computed as ($l$ is ignored here for simplicity):
\begin{equation}\label{eq:alpha}
\begin{aligned}
e_{ij} &= (\bm{h}_i\bm{W}_1 + \bm{h}_j\bm{W}_2 + p^{trans}_{ij} \bm{W}_3)\bm{W}_4^T, \\
\alpha_{ij} &= \frac{\exp({\rm LeakyReLU}(e_{ij}))}{\sum_{k \in \mathcal{N}_i}\exp({\rm LeakyReLU}(e_{ik}))},
\end{aligned}
\end{equation}
where $\bm{h}_i, \bm{h}_j\in\mathbb{R}^{d_l}$ are road representations of $v_i$ and $v_j$, $\bm{W}_1,\bm{W}_2\in\mathbb{R}^{d_{l} \times d_{l+1}},\bm{W}_3, \bm{W}_4\in\mathbb{R}^{1 \times d_{l+1}}$ are learnable parameters, ${\rm LeakyReLU}$ is the activation function whose negative input slope is 0.2~\cite{velivckovic2017graph}, and $p^{trans}_{ij}$ is the transfer probability between $v_i$ and $v_j$, which can be calculated as:
\begin{equation}
p^{trans}_{ij} = {\rm count}(v_i \rightarrow v_j) / {\rm count}(v_i),
\end{equation}
where ${\rm count}(v_i \rightarrow v_j)$ and ${\rm count}(v_i)$ is the frequency of edges $(v_i, v_j)$ and road $v_i$ appeared in the trajectory dataset $\mathcal{D}$, respectively.

Then we obtain the output feature $\tilde{\bm{h}}_i$ of the $i$-th road $v_i$ through combining the features of its neighborhoods using the attention weights as:
\begin{equation}\label{eq:sum}
\tilde{\bm{h}}_i^{(l+1)} = {\rm ELU}\left(\sum\limits_{j \in \mathcal{N}_i} \alpha_{ij} \bm{h}_j^{(l)}\bm{W}_5\right),
\end{equation}
where $\bm{W}_5$ are the learnable parameters and ${\rm ELU}$ is the Exponential Linear Unit activation function~\cite{velivckovic2017graph}. 

We use multi-head attention to stabilize the learning process and incorporate various types of information. Specifically, $H_1$ denotes the number of independent attention mechanisms that are computed as Equations \eqref{eq:alpha} and \eqref{eq:sum}, then we concatenate the outputs of these independent attention mechanisms as:
\begin{equation}\label{eq:multihead_att}
\bm{h}_i^{(l+1)} = \mathop{||}\limits_{k=1}^{H_1} {\rm ELU}\left(\sum\limits_{j \in \mathcal{N}_i} \alpha_{ij}^{(k)} \bm{h}_j^{(l)}\bm{W}_5^{(k)}\right),
\end{equation}
where $||$ represents concatenation, $\alpha_{ij}^{(k)}$ are the attention scores computed by the $k$-th attention head, $\bm{W}_5^{(k)}$ is the weight matrix of the corresponding linear transformation in layer $l$. 

The \gat layer considers the connectivity between roads due to both static road network structure and human mobility. The output of the last layer is defined as $\bm{r}_i \in \mathbb{R}^{d}$ and represents the representation of the road $v_i$, which contains road network contextual information and trajectory travel semantics. Moreover, the \gat layer is trained together with the trajectory encoder layer described below. We use sparse matrix operations following~\cite{velivckovic2017graph} to enable the model for large-scale road networks.

\subsection{Time-Aware Trajectory Encoder Layer}\label{ref:tatten}

After obtaining the road representations from the \gat layer, we need to convert road representation sequences into trajectory representations and incorporate temporal regularity information in the second stage. To model the co-occurrence relationship between roads in the trajectory, we use the Transformer encoder~\cite{vaswani2017attention} because it can capture the contextual information of the trajectory from the left and right sides of the road to realize the full interaction between roads. In addition, we extend the Transformer encoder and propose a Time-Aware Trajectory Encoder Layer (\enc) to incorporate temporal regularities in urban trajectories, which consist of two modules. The first is a Trajectory Time Pattern Extraction module that uses two temporal embeddings to capture the periodic patterns of urban traffic. The second is a Time Interval-Aware Self-Attention module to explicitly model the irregular time intervals between roads in the trajectory.

\subsubsection{\changed{Trajectory Time Pattern Extraction Module}} \changed{To capture the cyclical patterns of urban traffic, we use two temporal embedding vectors to extract the periodicity of weeks and days, respectively. For each visit timestamp $t_i$ of the road $v_i$, we use embedding vectors $\bm{t}_{mi{(t_i)}} \in \mathbb{R}^{d}$ and $\bm{t}_{di{(t_i)}} \in \mathbb{R}^{d}$ to embed the two periodic patterns, where ${mi{(t_i)}}$ and ${di{(t_i)}}$ are functions of transforming ${t}_i$ into its minutes index (1 to 1440) and day-of-week index (1 to 7).}

Then we obtain the fused embeddings $\bm{x}_i \in \mathbb{R}^{d}$ of road $v_i$ by summing several representations as follows:
\begin{equation}\label{eq:fusion}
\bm{x}_i = \bm{r}_i + \bm{t}_{mi{(t_i)}} + \bm{t}_{di{(t_i)}} + \bm{pe}_i,
\end{equation}
where $\bm{r}_i$ denotes the road representations, $\bm{t}_{mi{(t_i)}}$ and $\bm{t}_{di{(t_i)}}$ are corresponding temporal representations, and $\bm{pe}_i$ denotes the position encoding used in Transformer to introduce position information of the input trajectory. Finally, the initial representation of the trajectory $\mathcal{T}$ is obtained by concatenating the embeddings of roads in it as $\bm{X} = \bm{x}_1 \Vert \ldots \Vert \bm{x}_{|\mathcal{T}|} \in \mathbb{R}^{|\mathcal{T}| \times d}$.


\subsubsection{\changed{Time Interval-aware Self-attention Module}}
In the standard multi-head self-attention of Transformer encoder, given the input trajectory representation $\bm{X}$, the $H_2$ attention heads transform $\bm{X}$ into the $H_2$ query matrixes $\bm{Q}_h=\bm{X} \bm{W}_h^Q$, key matrixes $\bm{K}_h=\bm{X} \bm{W}_h^K$, and value matrixes $\bm{V}_h=\bm{X}\bm{W}_h^V$ synchronously, where $\bm{W}_h^Q, \bm{W}_h^K, \bm{W}_h^V \in \mathbb{R}^{d \times d'}$ are learnable parameters and $d' = d / H_2$. Then the self-attention of the $h$-th attention head is calculated as:
\begin{equation}\label{eq:self_att}
A_h(\bm{Q}_h, \bm{K}_h, \bm{V}_h) = {\rm softmax}\left(\frac{\bm{Q}_h\bm{K}_h^T}{\sqrt{d'}}\right)\bm{V}_h.
\end{equation}

\changed{To consider the irregular time intervals between road segments, which can reflect the congestion level of the road, we propose a \emph{Time Interval-Aware Self-Attention} to replace the standard self-attention of the Transformer encoder as:}
\begin{equation}\label{eq:timeaware_self_att}
TA_h(\bm{Q}_h, \bm{K}_h, \bm{V}_h) = {\rm softmax}\left(\frac{\bm{Q}_h\bm{K}_h^T}{\sqrt{d'}} + \tilde{\bm{\Delta}}\right)\bm{V}_h,
\end{equation}
\changed{where $\tilde{\bm{\Delta}}\in \mathbb{R}^{|\mathcal{T}| \times |\mathcal{T}|}$ is an adaptive time interval matrix and each element in it measures the impact among road segments in a trajectory. Given two roads $v_i$ and $v_j$, the element $\delta_{ij} \in \tilde{\bm{\Delta}}$ should have a large value when the time interval between $v_i$ and $v_j$ is short, \ie the two roads have strong impacts in the self-attention, vice versa. In this way, the irregular time intervals could be incorporated into the Transformer encoder.}

The calculation process of $\tilde{\bm{\Delta}}$ is as follows. Given the timestamp $t_i$ of the road $v_i$, we calculate the relative time interval $\delta _{i,j} = |t_i - t_j|$ for any two roads to obtain the original time interval matrix ${\bm{\Delta}}$ as:
\begin{equation}\label{eq:timeinterval_matrix}
{\bm{\Delta}} = 
\begin{bmatrix} \delta_{1,1} & \delta_{1,2} & \cdots & \delta_{1,|\mathcal{T}|} \\ \delta_{2,1} & \delta_{2,2} & \cdots & \delta_{2,|\mathcal{T}|} \\ \cdots & \cdots & \cdots & \cdots \\ \delta_{|\mathcal{T}|,1} & \delta_{|\mathcal{T}|,2} & \cdots & \delta_{|\mathcal{T}|,|\mathcal{T}|}
\end{bmatrix}.
\end{equation}

\changed{In the matrix ${\bm{\Delta}}$, the shorter the time interval between $v_i$ and $v_j$, the smaller the value of $\delta_{i,j}$. Since the impact between roads should become smaller with the time interval increasing, \ie the greater the time interval, the smaller the impact, we introduce a decay function to process the raw value in ${\bm{\Delta}}$. Specifically, we set $\delta_{i,j}' = {1}/{\log({\rm e} + \delta_{i,j})}$, where $e\approx2.718$. In this way, $\delta_{i,j}'$ decreases with increasing time intervals.}

\changed{Furthermore, we adopt a two-linear-transformation to process $\delta_{i,j}'$ as:}
\begin{equation}\label{eq:learnable_timeinterval}
\changed{\tilde{\delta}_{i,j} = ({\rm LeakyReLU}({\delta}_{i,j}'\, \bm{\omega}_1))\,\bm{\omega}_2^T,}
\end{equation}
\changed{where $\bm{\omega}_1, \bm{\omega}_2$ are learnable parameters and ${\rm LeakyReLU}$ is a activation function whose negative input slope is 0.2. By this method, $\tilde{\delta}_{i,j}$ becomes learnable and can capture the irregular time interval information. Finally, we plug $\tilde{\delta}_{i,j}$ into Eq.~\eqref{eq:timeaware_self_att} to get the Time Interval-Aware Self-Attention.}

Then we concatenate the output of the $H_2$ attention heads and project it through $\bm{W}^O \in \mathbb{R}^{d \times d}$ to obtain the outputs $\bm{X}' \in \mathbb{R}^{|\mathcal{T}| \times d}$ as:
\begin{equation}\label{eq:timeaware_multihead_att}
\bm{X}' = {\rm MultiAtt}(\bm{Q}, \bm{K}, \bm{V}) = (TA_1 \Vert \ldots \Vert {TA_{H_2}})\bm{W}^O.
\end{equation}

After the multi-head attention, we employ layer normalization and residual connection following Transformer~\cite{vaswani2017attention}. \changed{Finally, a position-wise feed-forward network (noted as ${\rm FFN}$) consists of two layers of linear transformations,} and ${\rm ReLU}$ activation is used to get the output representation $\bm{Z}$ of trajectory $\mathcal{T}$ as:
\begin{equation}\label{eq:feed_forwared}
\bm{Z} = ({\rm ReLU} (\bm{X}' \bm{W}_{F}^1 + \bm{b}_{F}^1)) \bm{W}_{F}^2 + \bm{b}_{F}^2,
\end{equation}
where $\bm{W}_{F}^1, \bm{W}_{F}^2 \in \mathbb{R}^{d \times d}, \bm{b}_{F}^2, \bm{b}_{F}^2 \in \mathbb{R}^{d}$ are learnable parameters and ${\rm ReLU}$ is the activation function. The layer normalization and residual connection are also used here. 


\subsubsection{Trajectory Representation Pooling}
After stacking $L_2$ layers of the self-attention module, we obtain the final output representation $\bm{Z} \in \mathbb{R}^{|\mathcal{T}| \times d}$, which has been fully interacted between the road segments. Furthermore, following ~\cite{devlin2018bert}, we extract the whole trajectory representation ${\bm{p}_i} \in \mathbb{R}^{d}$ by inserting a placeholder in the first position throughout training tasks and take it as the trajectory representation.



\subsection{Self-supervised Pre-training Tasks}
This work aims to learn trajectory representations self-supervised to support multiple downstream tasks. Therefore, considering the spatial-temporal characteristics of the trajectories, we design two self-supervised tasks which do not target specific downstream tasks to learn generic representations.

\subsubsection{Span-Masked Trajectory Recovery}
Masked language modeling (MLM) has proven its superiority in learning sequence data representations in many studies~\cite{devlin2018bert, chen2021robust}. Each word in the sequence is masked independently with a probability in the previous MLM task, and the model is used to predict the masked words. However, this task is not fully applicable to our task because the trajectory is a sequence of \textit{adjacent} roads. If we mask the road independently, the model can easily infer the masked road based on its upstream and downstream roads in the road network. Therefore, we propose the span-masked method, where we select several consecutive subsequences of length $l_m$ in the trajectory for masking, whose total length is $p_{m}$ percent of the trajectory length. \changed{When masking the trajectory, we replace the selected road $v_i$ with a special token ${\rm [MASK]}$ and set the corresponding minute index ${mi{(t_i)}}$ and day-of-week index ${di{(t_i)}}$ to a special token ${\rm [MASKT]}$.} After obtaining the representation $\bm{Z}$ of the masked trajectory $\mathcal{T}$, we use a linear layer with parameters $\bm{W}_{m} \in \mathbb{R}^ {d \times {|\mathcal{V}|}}, \bm{b}_{m} \in \mathbb{R}^{|\mathcal{V}|}$ to predict the masked roads as:
\begin{equation}
    \bm{\hat{Z}} = \bm{Z}\bm{W}_{m} + \bm{b}_{m} \in \mathbb{R}^ {|\mathcal{T}| \times {|\mathcal{V}|}},
\end{equation}
Then we use the cross-entropy loss between masked roads and predicted values as the optimized target:
\begin{equation}
    \bm{\mathcal{L}}^{mask}_{\mathcal{T}} = -\frac{1}{|\mathcal{M}|}\sum_{v_i \in \mathcal{M}} \log \frac{\exp(\bm{\hat{Z}}_{v_i})}{\sum_{v_j \in \mathcal{V}}\exp(\bm{\hat{Z}}_{v_j})},
\end{equation}
where $\mathcal{M}$ is the set of masked roads. We average all losses of $N_{b}$ trajectories in a mini-batch to obtain the loss $\bm{\mathcal{L}}^{mask}$.

\subsubsection{Trajectory Contrastive Learning} 
Mask prediction focuses on capturing co-occurrence relationships between roads and contextual information of the road network. To improve the modeling of the spatial-temporal characteristics and travel semantics, we introduce a contrastive learning method.

\emph{Trajectory Data Augmentation Strategies.} Contrastive learning aims to learn representations to bring semantically similar positive samples closer and make negative samples farther apart. Thus, the crucial question is how to construct different views in contrastive learning. Considering the spatial-temporal characteristics of the trajectories, we explore four data augmentation strategies to generate views for contrastive learning.

\begin{itemize}
  \item \emph{Trajectory Trimming}: We obtain the enhanced trajectory by randomly removing a continuous subsequence from the trajectory. In order not to destroy the continuity and travel semantics of the trajectory, we trim only at the origin or destination of the trajectory, and the trimming ratio $r_1$ is a random sample of $0.05-0.15$. This data augmentation method is applied since the semantics of trajectories with close origins or destinations are similar.
  \item \emph{Temporal Shifting}: Influenced by the urban traffic patterns, the road travel time is dynamic. Given a trajectory, we randomly select a subset of roads (scale $r_2 = 0.15$) and perform a random perturbation by $t_{aug} = t_{cur} - (t_{cur} - t_{his}) * r_3$, where $r_3$ is a random sample of $0.15-0.30$, $t_{cur}$ and $t_{his}$ are the current and historical average travel time of that road, respectively. Using this augmentation method helps to capture the travel semantics of the trajectory in the temporal dimension.
  \item \emph{Road Segments Mask}: In the span-masked trajectory recovery task, some roads and the corresponding timestamps of the trajectory are randomly selected and masked. The masked trajectory can be considered as the trajectory with missing values to learn the travel semantics of the trajectories in both temporal and spatial dimensions.
  \item \emph{Dropout}: Dropout is a widely used method to avoid overfitting. Here we use it as a data augmentation method to randomly drop some tokens with a certain probability from the data embedding layer and set them to zero~\cite{gao2021simcse}. 
\end{itemize}

\emph{Contrastive Trajectory Learning.} Following~\cite{chen2020simple}, we adopt the normalized temperature-scaled cross-entropy loss with in-batch negatives as the contrastive objective. We randomly select $N_{b}$ trajectories from the dataset $\mathcal{D}$ and then obtain $2N_{b}$ trajectories after data augmentation. Each trajectory (also called the anchor) is trained to find out the corresponding data-augmentation trajectory (the positive sample) among $2(N_{b}-1)$ negative samples in the batch. Formally, the contrastive training objective for a positive pair $(i, j)$ is defined as:
\begin{equation}
    \bm{\mathcal{L}}^{con}_{i,j} = -\log \frac{\exp({\rm sim}(\bm{p}_i, \bm{p}_j)/\tau)}{\sum_{k=1}^{2N}\textbf{1}_{[k \ne i]}\exp({\rm sim}(\bm{p}_i, \bm{p}_k)/\tau)},
\end{equation}
where $\tau$ is the temperature hyperparameter, \changed{${\rm sim}({\bm{p}_i}, \bm{p}_j)$ is the cosine similarity $\frac{{\bm{p}_i} \cdot \bm{p}_j}{\Vert {\bm{p}_i}\Vert \Vert\bm{p}_j\Vert}$ between ${\bm{p}_i}$ and ${\bm{p}_j}$ ($\cdot$ is the inner product operation)}, $\textbf{1}$ is the indicator equal to one if the condition is satisfied, otherwise it is zero. We average all $2N_{b}$ in-batch losses to obtain the contrastive loss $\bm{\mathcal{L}}^{con}$.



\changed{We pre-train the proposed \name with the two self-supervised tasks above. The pre-training loss is defined as:
\begin{equation}
    \bm{\mathcal{L}}^{pre} = \lambda \bm{\mathcal{L}}^{mask} + (1 - \lambda) \bm{\mathcal{L}}^{con},
\end{equation}
where $\lambda$ is the hyperparameter to balance the two tasks.}

\vspace{-.1cm}
\subsection{Model Fine-tuning and Downstream Tasks}
In this section, we aim to adapt the learned representations to specific downstream tasks, either directly or with the necessary fine-tuning.

\subsubsection{Trajectory Travel Time Estimation}

This task aims to estimate the travel time from the origin to the destination with a given road sequence and the departure time. We build a regression model using a single fully connected layer to obtain the predicted value as $\hat{y}_i = FC ({\bm{p}_i})$. Then we use the mean square error (MSE) as the optimization objective:
\begin{equation}\label{exp:reg} \bm{\mathcal{L}}^{regress} = \frac{1}{N} \sum_{i=1}^N \Vert y_i - \hat{y}_i \Vert ^2,
\end{equation}
where $y_i$ is the ground truth and $N$ is the total number of trajectories in the test dataset. 

\subsubsection{Trajectory Classification}
This task aims to classify trajectories based on a specific label, such as carrying passengers or not, the driver ID, the transportation, etc. We employ a simple fully connected layer with the ${\rm softmax}$ activation to obtain the predicted value as $\bm{\hat{y}}_i = {\rm softmax}(FC ({\bm{p}_i}))$. Then we optimize the model with the cross-entropy loss:
\begin{equation}\label{exp:classify} \bm{\mathcal{L}}^{classify} = \frac{1}{N} \sum_{i=1}^N \sum_{c=1}^C -\bm{{y}}_i(c)\log(\bm{\hat{y}}_i(c)),
\end{equation}
where $\bm{{y}}_i$ is the ground truth, $N$ is the total number of trajectories in the test dataset, and $C$ is the number of categories.

\subsubsection{Trajectory Similarity Computation and Search}
In this task, we design two sub-tasks: the most similar trajectory search and the $k$-nearest trajectory search. Here we directly use the representation ${\bm{p}_i}$ obtained from the pre-training task without fine-tuning. The most similar trajectory search task is to find out the most similar trajectory from a large database given a query. In the $k$-nearest trajectory search task, given a trajectory, models need to find top-$k$ similar trajectories from candidates ignoring the rank. The detailed settings for these two tasks can be found in Section~\ref{exp:sim}.

\makeatletter
\newcommand{\thickhline}{%
    \noalign {\ifnum 0=`}\fi \hrule height 1pt
    \futurelet \reserved@a \@xhline
}

\section{Experiments}

\changed{In this section, we conduct extensive experiments to evaluate the performance of the \name framework. The experiments include five parts:}
\begin{itemize}
  \item \changed{{\em Performance Comparison}. We compare the performance of \name with eight baselines on two large-scale datasets for three downstream tasks. The experiment results show the superior overall performance of \name.}
  \item \changed{{\em Pre-training Effect Study}. We demonstrate the effectiveness of the self-supervised pre-training tasks over small-size datasets and across datasets. The results show that the self-supervised tasks can effectively reduce the usage of training data, and the model can be transferred across heterogeneous datasets. This nature is beneficial for solving the problem of insufficient training data.}
  \item \changed{{\em Ablation Experiment}. We use ablation studies to verify the effectiveness of each sub-module of \name.}
  \item \changed{{\em Parameter Sensitivity Experiment}. This experiment verifies the stability of our method over key parameters.}
  \item \changed{{\em Efficiency and Scalability Study}. This experiment clarifies that our proposed framework is efficient and can scale for large datasets.}
\end{itemize}

\subsection{Datasets and Preprocessing}\label{exp:dataset}
We use two real-world, large-scale trajectory datasets in the experiments, \ie \bj and \porto. \bj was collected by taxis in Beijing in November 2015. \porto is an open-source dataset released for a Kaggle competition\footnote{\url{https://www.kaggle.com/c/pkdd-15-predict-taxi-service-trajectory-i}} and is sampled every 15 seconds. \changed{We download map data of Beijing and Porto from OpenStreetMap (OSM)~\cite{OpenStreetMap} to construct the directed graph (road network) $\mathcal{G} = (\mathcal{V}, \mathcal{E}, \bm{F}_\mathcal{V}, \bm{A})$ defined in Definition 1. The OSM data contains three parts: $i$) All road segments in the cities. We use these roads to form the vertex set $\mathcal{V}$ of $\mathcal{G}$, where each road represents a vertex. $ii$) The connection relationships between all roads. If two roads have a connection relation, we define that the corresponding vertexes have an edge. We use these connection relationships to form the binary value adjacency matrix $\bm{A}$ and the edge set $\mathcal{E}$ of $\mathcal{G}$. $iii$) The features of roads. We select four important road features, \ie road type, length, number of lanes, and maximum travel speed, and calculate the in-degree and out-degree of each road in the adjacency matrix $\bm{A}$ to construct the road features $\bm{F}_\mathcal{V}$. Finally, we use the directed graph $\mathcal{G}$ of Beijing and Porto as the input of our proposed \name.} Furthermore, we perform \textit{map matching}~\cite{yang2018fast} to obtain the road-network constrained trajectories. Details of the two datasets are given in Table~\ref{tab:data_detail}. 


We ignore the roads that are not covered by the trajectories. Besides, we also remove loop trajectories, trajectories with lengths less than six, and users with less than 20 trajectories and set the maximum trajectory length to 128. We split \bj into the training, validation, and test datasets in chronological order. The three datasets cover 18/5/7 days because there is less data on November 25. For \porto, we split each month's data in chronological order with a ratio of 6:2:2 and combine the data per month in the training, validation, and test datasets, considering the effects of seasons. We use the same data partitioning method in the pre-training and fine-tuning phases. The codes and processed datasets are available here~\footnote{\url{https://github.com/aptx1231/START}}.


\begin{table}[t]
  \centering
  \caption{Statistics of the two datasets after preprocessing.}
  \resizebox{\columnwidth}{!}{
    \begin{tabular}{c|c|c}
    \thickhline
    Dataset & \bj  & \porto \\
    \hline
    Time span & 2015/11/01-2015/11/30 & 2013/07/01-2014/07/01 \\
    \hline
    \#Trajectory & 1018312 & 695085 \\
    \hline
    \#Usr & 1677  & 435 \\
    \hline
    \#Road Segment & 38479 & 10903 \\
    \hline
    train/eval/test  & 656221/174478/187613 & 417040/139020/139025 \\
    \thickhline
    \end{tabular}%
    }
  \label{tab:data_detail}%
\vspace{-.5cm}
\end{table}%

\begin{table*}[htbp]
  \centering
  \caption{Three Downstream Tasks Overall Performance on \bj and \porto.}
  \resizebox{\textwidth}{!}{
  \begin{threeparttable}
    \begin{tabular}{c|c|ccc|ccc|ccc}
    \thickhline
     \multicolumn{2}{c|}{} & \multicolumn{3}{c|}{Travel Time Estimation} & \multicolumn{3}{c|}{Trajectory Classification} & \multicolumn{3}{c}{Most Similar Trajectory Search}  \\
    \thickhline
    \multirow{12}{*}{{\rotatebox{90}{\bj}}} & Models & \multicolumn{1}{c}{MAE $\downarrow$} & \multicolumn{1}{c}{MAPE(\%) $\downarrow$} & RMSE $\downarrow$  & ACC $\uparrow$  & F1 $\uparrow$    & AUC $\uparrow$   & MR $\downarrow$   & HR@1 $\uparrow$  & HR@5 $\uparrow$ \\
\cline{2-11}                   & traj2vec & 10.13{$\pm$}0.12  & 37.95{$\pm$}0.92  & 56.83{$\pm$}0.47  & 0.811{$\pm$}6e-4  & 0.852{$\pm$}1e-3  & 0.873{$\pm$}2e-5  & 7.186{$\pm$}0.03  & 0.607{$\pm$}1e-4  & 0.766{$\pm$}7e-5   \\
\cline{2-11}                   & t2vec & 10.03{$\pm$}0.10  & 36.42{$\pm$}1.31  & 56.65{$\pm$}0.12  & 0.814{$\pm$}1e-3  & 0.863{$\pm$}1e-2  & 0.879{$\pm$}9e-4  & 5.948{$\pm$}0.01  & 0.788{$\pm$}3e-4  & 0.935{$\pm$}8e-5   \\
\cline{2-11}                   & Trembr & \underline{9.997}{$\pm$}0.11  & \underline{34.20}{$\pm$}0.88  & \underline{36.97}{$\pm$}0.38  &  \underline{0.818}{$\pm$}1e-3  & \underline{0.871}{$\pm$}2e-3  & \underline{0.880}{$\pm$}2e-3  & \underline{2.509}{$\pm$}2e-3  & \underline{0.884}{$\pm$}5e-4  & \underline{0.952}{$\pm$}6e-5   \\
\cline{2-11}                   & Transformer & 10.74{$\pm$}0.48  & 39.61{$\pm$}1.52  & 57.16{$\pm$}0.56  & 0.794{$\pm$}2e-3  & 0.845{$\pm$}1e-3  & 0.846{$\pm$}1e-3  & 40.60{$\pm$}0.19  & 0.515{$\pm$}1e-4  & 0.649{$\pm$}1e-4   \\
\cline{2-11}                   & BERT  & 10.21{$\pm$}0.14  & 37.31{$\pm$}0.17  & 37.09{$\pm$}0.36  & 0.804{$\pm$}3e-3  & 0.862{$\pm$}2e-3  & 0.864{$\pm$}3e-3  & 27.10{$\pm$}0.11  & 0.587{$\pm$}3e-3  & 0.712{$\pm$}4e-4  \\
\cline{2-11}                   & PIM   & 10.19{$\pm$}0.09  & 39.04{$\pm$}0.58  & 57.73{$\pm$}0.26  & 0.803{$\pm$}1e-3  & 0.861{$\pm$}1e-3  & 0.862{$\pm$}9e-4  & 23.51{$\pm$}0.12  & 0.760{$\pm$}4e-3  & 0.898{$\pm$}2e-4   \\
\cline{2-11}                   & PIM-TF & 12.05{$\pm$}0.03  & 43.14{$\pm$}0.66  & 61.15{$\pm$}0.33  & 0.789{$\pm$}1e-3  & 0.849{$\pm$}2e-3  & 0.842{$\pm$}6e-3  & 86.45{$\pm$}0.32  & 0.296{$\pm$}2e-4  & 0.340{$\pm$}2e-4   \\
\cline{2-11}                   & Toast & 10.69{$\pm$}0.22  & 35.37{$\pm$}1.14  & 57.41{$\pm$}0.41  & 0.810{$\pm$}2e-3  & 0.870{$\pm$}2e-3  & 0.871{$\pm$}2e-3  & 29.53{$\pm$}0.15  & 0.611{$\pm$}2e-4  & 0.746{$\pm$}3e-4   \\
\cline{2-11}                   & \name & \textbf{9.134}{$\pm$}0.03 & \textbf{30.92}{$\pm$}0.35 & \textbf{35.40}{$\pm$}0.09 & \textbf{0.853}{$\pm$}2e-3 & \textbf{0.896}{$\pm$}1e-3 & \textbf{0.916}{$\pm$}4e-4 & 
\textbf{1.295}{$\pm$}1e-3 & \textbf{0.969}{$\pm$}4e-4 & \textbf{0.997}{$\pm$}4e-5  \\
\cline{2-11}                   & Improve & 8.63\% & 9.59\% & 4.24\% & 4.28\% & 2.87\% & 4.09\% & 48.39\% & 9.62\% & 4.73\%  \\
    \thickhline
    \multirow{12}{*}{{\rotatebox{90}{\porto}}} & Models & MAE $\downarrow$   & MAPE $\downarrow$  & RMSE $\downarrow$  & Micro-F1 $\uparrow$ & Macro-F1 $\uparrow$ & Recall@5 $\uparrow$ & MR $\downarrow$    & HR@1 $\uparrow$  & HR@5  $\uparrow$ \\
\cline{2-11}          & traj2vec & 1.552{$\pm$}6e-3  & 23.70{$\pm$}0.35  & 2.351{$\pm$}4e-3  & 0.063{$\pm$}3e-3  & 0.038{$\pm$}3e-3  & 0.183{$\pm$}5e-3  & 30.52{$\pm$}0.13  & 0.552{$\pm$}3e-4  & 0.732{$\pm$}5e-4  \\
\cline{2-11}          & t2vec & 1.539{$\pm$}5e-3  & 23.65{$\pm$}0.12  & 2.324{$\pm$}5e-3  & 0.068{$\pm$}2e-4  & 0.048{$\pm$}3e-4  & 0.187{$\pm$}3e-4  & 12.70{$\pm$}0.08  & 0.746{$\pm$}3e-4  & 0.856{$\pm$}8e-4  \\
\cline{2-11}          & Trembr & \underline{1.480}{$\pm$}2e-3  & \underline{22.64}{$\pm$}0.37  & \underline{2.164}{$\pm$}0.01  & \underline{0.071}{$\pm$}9e-4  & \underline{0.049}{$\pm$}1e-3  & \underline{0.192}{$\pm$}2e-3  & \underline{4.635}{$\pm$}1e-3  & \underline{0.846}{$\pm$}4e-4  & \underline{0.929}{$\pm$}8e-5  \\
\cline{2-11}          & Transformer & 1.738{$\pm$}3e-3  & 25.72{$\pm$}0.26  & 2.637{$\pm$}2e-3  & 0.028{$\pm$}7e-3  & 0.018{$\pm$}5e-3  & 0.075{$\pm$}8e-3  & 68.58{$\pm$}0.21  & 0.447{$\pm$}2e-4  & 0.664{$\pm$}5e-5  \\
\cline{2-11}          & BERT  & 1.593{$\pm$}7e-3  & 24.63{$\pm$}0.57  & 2.291{$\pm$}3e-3  & 0.065{$\pm$}3e-4  & 0.044{$\pm$}1e-3  & 0.184{$\pm$}1e-3  & 39.12{$\pm$}0.15  & 0.511{$\pm$}4e-3  & 0.714{$\pm$}5e-4  \\
\cline{2-11}          & PIM   & 1.559{$\pm$}3e-3  & 24.68{$\pm$}0.25  & 2.339{$\pm$}0.01  & 0.061{$\pm$}4e-4  & 0.037{$\pm$}3e-4  & 0.153{$\pm$}5e-4  & 19.53{$\pm$}0.10  & 0.653{$\pm$}3e-4  & 0.774{$\pm$}7e-4  \\
\cline{2-11}          & PIM-TF & 1.945{$\pm$}2e-3  & 28.82{$\pm$}0.15  & 2.841{$\pm$}3e-4  & 0.025{$\pm$}4e-3  & 0.016{$\pm$}5e-3  & 0.069{$\pm$}7e-3  & 78.78{$\pm$}0.24  & 0.384{$\pm$}2e-5  & 0.547{$\pm$}3e-5  \\
\cline{2-11}          & Toast & 1.624{$\pm$}8e-3  & 24.63{$\pm$}0.33  & 2.445{$\pm$}5e-3  & 0.062{$\pm$}1e-3  & 0.035{$\pm$}4e-4  & 0.181{$\pm$}1e-3  & 22.61{$\pm$}0.12  & 0.684{$\pm$}2e-5  & 0.789{$\pm$}2e-5  \\
\cline{2-11}          & \name & \textbf{1.334}{$\pm$}3e-3 & \textbf{20.66}{$\pm$}0.14 & \textbf{2.001}{$\pm$}1e-3 & \textbf{0.089}{$\pm$}4e-4 & \textbf{0.067}{$\pm$}2e-3 & \textbf{0.244}{$\pm$}1e-3 & \textbf{1.897}{$\pm$}1e-3 & \textbf{0.921}{$\pm$}3e-4 & \textbf{0.973}{$\pm$}6e-5 \\
\cline{2-11}          & Improve & 9.86\% & 8.75\% & 7.53\% & 25.35\% & 36.73\% & 27.08\% & 59.07\% & 8.87\% & 4.74\% \\
    \thickhline
    \end{tabular}%
    \begin{tablenotes}
        \footnotesize
        \item[*] All experiments are repeated ten times, and we report both the mean and standard deviation. The bold results are the best, and the underlined results are the second best. The metric with "$\uparrow$" means that a larger result is better, and the metric "$\downarrow$" means that a smaller result is better.  
      \end{tablenotes}
    \end{threeparttable}
    }
  \label{exp:tableresult}%
  \vspace{-0.6cm}
\end{table*}%

\subsection{Baselines}\label{exp:baseline}

\changed{We select the trajectory representation learning methods that {\em adopt self-supervised training methods} and are {\em suitable for multiple downstream tasks}, \ie non-task-specific methods, as our baselines. The baselines meet the criteria include three categories: }

\changed{(1) \emph{Encoder-decoder with reconstruction}: This category uses an RNN-based encoder-decoder model to convert raw trajectories as representation vectors and adopts the reconstruction self-supervised task to train the encoder-decoder model. We select the following representative methods as the baselines.}
\begin{itemize}
    \item Traj2vec~\cite{yao2017trajectory} converts trajectories to feature sequences and uses a sequence-to-sequence (seq2seq) model to learn representations.
    \item T2vec~\cite{li2018deep} is the state-of-the-art seq2seq trajectory representation method with negative sampling and spatial proximity aware loss. We use the decoder of t2vec to recover the input trajectory without downsampling since the data are road-network constrained trajectories.
    \item Trembr~\cite{fu2020trembr} is a seq2seq model whose decoder reconstructs both roads and timestamps of the input trajectory.
\end{itemize}

\changed{(2) \emph{Two-stage representation models}: This category first converts road segments as representation vectors and then generates trajectory representation vectors from the road representation vectors in the same trajectory. We select the following methods of this category as the baselines.}
\begin{itemize}
    \item PIM~\cite{yang2021unsupervised} uses node2vec to generate road representations of the static road network and uses a mutual information maximization method to train a LSTM encoder for trajectory representation generation. 
    \item PIM-TF replaces the LSTM encoder in PIM with a transformer encoder.
    \item Toast~\cite{chen2021robust} uses the context-aware node2vec to generate road representations and uses the MLM and trajectory discrimination task to train a Transformer encoder for trajectory representation generation.
\end{itemize}

\changed{Besides, we also adopt classical self-supervised sequence representation learning models as the baselines.}

\changed{(3) \emph{Self-supervised sequence representation models}: We adopt Transformer and BERT that input with road-network constrained trajectories as the baselines.}
\begin{itemize}
    \item Transformer~\cite{vaswani2017attention} is a self-attention model of the encoder-decoder architecture. We use MLM as the pre-training self-supervised task.
    \item BERT~\cite{devlin2018bert} is a self-attention model. We train the model with a MLM task and a classification task, splitting a trajectory $\mathcal{T}$ as two parts, \ie $\mathcal{T}_1$ and $\mathcal{T}_2$, and treating $(\mathcal{T}_1, \mathcal{T}_2)$ as positive samples and $(\mathcal{T}_2, \mathcal{T}_1)$ as negative samples.
\end{itemize}

\changed{The models mentioned in the related work section but requiring supervised labels, such as NEUTRAJ~\cite{yao2019computing}, Traj2SimVec~\cite{zhang2020trajectory} and T3S~\cite{yang2021t3s}, and the methods that are not suitable for multiple downstream tasks, such as DETECT~\cite{yue2019detect} and E2dtc~\cite{fang20212}, are not chosen as the baselines.}

\subsection{Experimental Settings}\label{exp:set}

\subsubsection{Model and Baseline Settings}
All experiments are conducted on Ubuntu 18.04 with an NVIDIA GeForce 3090 GPU. We implement \name and all baselines based on the PyTorch 1.7.1~\cite{pytorch}. We set the embedding size $d$ to 256, the \gat layers $L_1$ to 3, and the \enc layers $L_2$ to 6. The attention heads $H_1$ are [8, 16, 1] for \gat and $H_2$ is 8 for \enc. The mask length $l_m$ is 2 and the mask ratio $p_{m}$ is 15\%. The dropout ratio is 0.1, and the temperature parameter $\tau$ is 0.05. The default data augmentation methods are \textit{Trajectory Trimming} and \textit{Temporal Shifting}. Finally, $\lambda=0.6$ to balance the pre-training losses. The baselines have the same settings as \name, with 256 hidden dimensions and six layers (or six encoders and six decoders for the encoder-decoder model), and the other settings follow their defaults.

\subsubsection{Training Settings} 
We pre-train and fine-tune our model using the optimizer AdamW~\cite{loshchilov2018fixing}. The batch size is 64, and the training epoch is 30. The learning rate $lr$ is 0.0002, and we use the warm-up policy corresponding to increase $lr$ linearly for the first five epochs and decrease it after using a cosine annealing schedule. 

\subsubsection{Evaluation Metrics}  
For the travel time estimation task, we adopt three metrics, including mean absolute error (MAE), mean absolute percentage error (MAPE), and root mean square error (RMSE). For the trajectory classification task, we use Accuracy (ACC), F1-score (F1), and Area Under ROC (AUC) to evaluate binary classification tasks, and Micro-F1, Macro-F1, and Recall@5 to evaluate multi-classification tasks. For the most similar search task, we use Mean Rank(MR) and Hit Ratio(HR@1, HR@5) to evaluate whether the model can find the truth. For the $k$-nearest search task, we use Precision to measure the coverage of the top-$k$ results. 

\subsection{\changed{Performance Comparision}}
\subsubsection{\changed{Overall Performance}}
\changed{Table~\ref{exp:tableresult} reports the overall results on the three downstream tasks. Based on the table, we can make the following observations.}

\begin{itemize}
 \item \changed{Our \name achieves the best performance in terms of all metrics on these three tasks for the two real-world datasets. It confirms the superior performance of our framework in learning trajectory representations by introducing temporal regularities and travel semantics in the pre-training phase.}
 \item \changed{The encoder-decoder models with reconstruction outperform the sequence representation models (Transformer, BERT). It could be because the pre-training methods of these two models from the natural language processing domain are not suitable for trajectory data, ignoring the spatial-temporal characteristics.}
 \item \changed{Trembr performs best among all baselines because it considers the visit timestamp of each road in the decoding process, highlighting the importance of temporal information in the trajectories.}
 \item \changed{The performance of the two-stage models, \ie PIM and Toast, is unsatisfactory due to two factors. First, they consider trajectories as ordinary road sequences and ignore the temporal information. Second, their road representation learning method does not adequately consider the travel semantics, such as road visit frequencies.}
\end{itemize}

\begin{figure}[t]
    \centering
    \includegraphics[width=0.95\columnwidth]{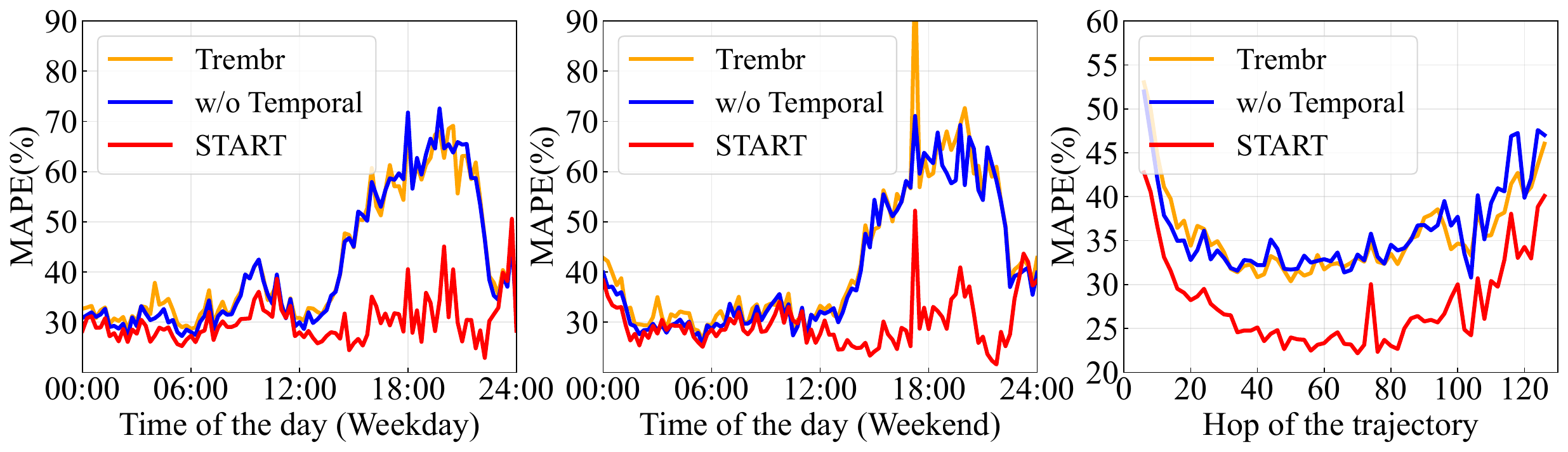}
    \vspace{-.2cm}
    \caption{MAPE on \bj Under Different Scenarios.}
    \label{fig:ETA}
    \vspace{-.4cm}
\end{figure}

\subsubsection{Performance of Trajectory Travel Time Estimation}\label{exp:eta}
We fine-tune all models with the objective function \eqref{exp:reg}. Note that no time information is fed into the model during fine-tuning, except for the \textit{departure time} to avoid information leakage. In addition to the overall performance in Table~\ref{exp:tableresult}, to investigate the performance of the model under different scenarios and verify the role of pre-training with temporal regularities, we present the MAPE results on different departure times, whether it is a weekend or not, and the hops of the trajectory on \bj in Figure~\ref{fig:ETA}. Here we compare three models, including \name, a variant without temporal (noted as \textit{w/o Temporal}) where the time embeddings and the time interval matrix are removed, and the best baseline Trembr. We can observe the following phenomena: (1) \name consistently outperforms others, regardless of the weekday or weekend or the trajectory hop size. Moreover, \name shows excellent performance, especially in the late peak periods (16:00-21:00) and when the trajectory is between 20 to 100 hops. (2) The no temporal variant cannot capture temporal regularities and therefore has worse performance than \name, highlighting the importance of temporal regularities when pre-training.

\subsubsection{Performance of Trajectory Classification} 
We fine-tune all models with the objective function \eqref{exp:classify}. We use whether the taxi carries passengers as a binary classification label in \bj and the driver ID as the label in \porto for the multi-classification (435 classes). As shown in Table~\ref{exp:tableresult}, our model consistently outperforms all baselines because it can capture the underlying travel semantics and achieves accurate performance.


\subsubsection{Performance of Trajectory Similarity Search} \label{exp:sim}

The similarity measure is a fundamental problem with various applications, such as identifying popular routes and similar drivers in trajectory analysis. A recent study~\cite{li2018deep} proposes to use the most similar trajectory search and the $k$-nearest trajectory search to evaluate the effectiveness of different methods. We adopt it in the experiments as it is currently the best evaluation method that proves the effectiveness of the model from multiple perspectives. Here we directly use the trajectory representations obtained from the pre-training without fine-tuning and use the Euclidean distance of the representations to represent the similarity between the trajectories, \ie the smaller the distance, the greater the similarity.

\textbf{\emph{(a) Most Similar Trajectory Search:}}
The most similar trajectory search task is to find out the most similar trajectory $\mathcal{T}_a'$ from a large trajectory database $\mathcal{D_D}$ given a query trajectory $\mathcal{T}_a$ in the query dataset $\mathcal{D_Q}$. However, the lack of ground truth makes it difficult to evaluate the accuracy of trajectory similarity. Li \etal~\cite{li2018deep} use downsampling in various proportions to construct the query and ground truth from the GPS-based trajectories. However, the influence of downsampling can be eliminated after the \textit{map matching}. Chen \etal~\cite{chen2021robust} propose a detour method to generate ground truth for road-network constrained trajectories. Based on this, we propose a ground truth generation method based on the top-$k$ detour. Specifically, we randomly select $N_{q}$ trajectories from the test dataset, denoted as the query dataset $\mathcal{D_Q}$. For each trajectory $\mathcal{T}_a \in \mathcal{D_Q}$, we select a section of consecutive sub-trajectories $\mathcal{S}_a$ whose length does not exceed $p_d$ (\eg 0.2) of the original trajectory length. Then we perform a top-$k$ search~\cite{yen1971finding} on the road network between the origin and destination of $\mathcal{S}_a$. If the travel time of the searched trajectory exceeds a certain threshold $t_d$ with respect to the original trajectory, this trajectory is defined as $\mathcal{S}_a'$. The detour trajectory $\mathcal{T}_a'$ of $\mathcal{T}_a$ is obtained by replacing $\mathcal{S}_a$ by $\mathcal{S}_a'$. In this way, we can construct the detour dataset $\mathcal{D_Q}' = \{\mathcal{T}_a'\}$. Furthermore, we extract other $N_{neg}$ trajectories from the test dataset that do not overlap with $\mathcal{D_Q}$, defined as the set $\mathcal{D_N}$, and use the same method to obtain the corresponding detour dataset $\mathcal{D_N}'$. Together, $\mathcal{D_N}'$ and $\mathcal{D_Q}'$ form the database $\mathcal{\mathcal{D_D}} = \mathcal{D_N}' \cup \mathcal{D_Q}'$. When using $\mathcal{T}_a$ to query the most similar trajectory in $\mathcal{D_D}$, $\mathcal{T}_a'$ will ideally rank first since it is generated from $\mathcal{T}_a$, \ie the ground truth of $\mathcal{T}_a$.

\begin{figure}[t]
\centering
\includegraphics[width=0.95\columnwidth]{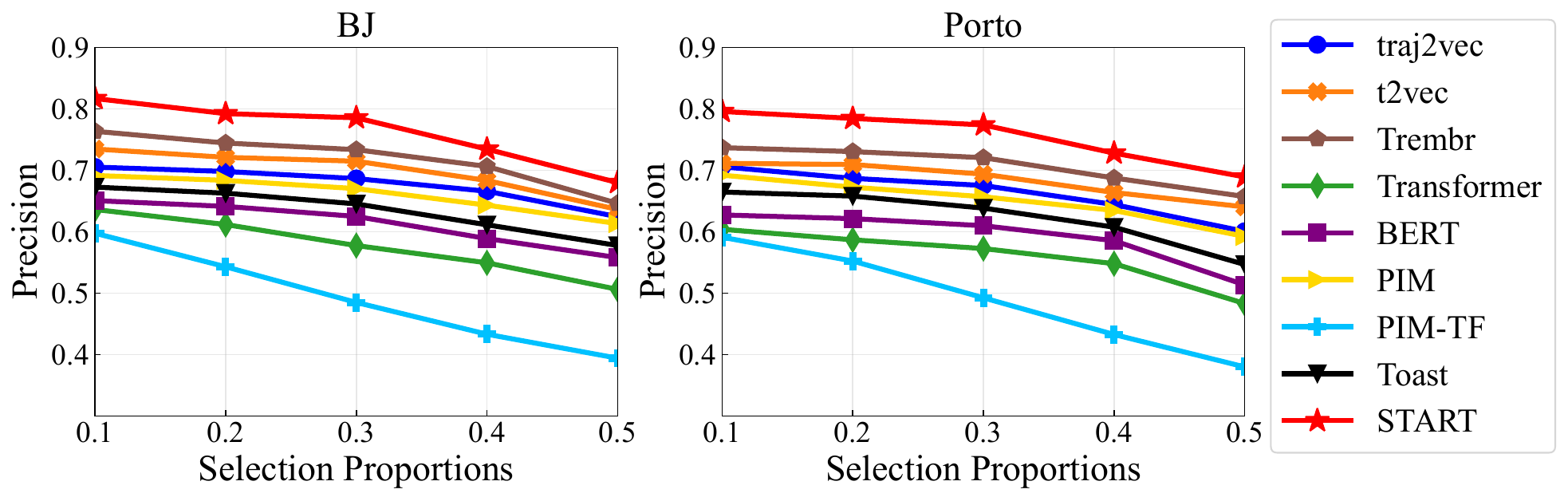}
    \vspace{-.3cm}
    \caption{Performance of $k$-nearest Trajectory Search Task When Selection Proportions $p_d$ Vary.}
    \label{fig:proportions}
    \vspace{-0.7cm}
\end{figure}

\begin{figure*}[htbp]
\centering
\includegraphics[width=0.95\textwidth]{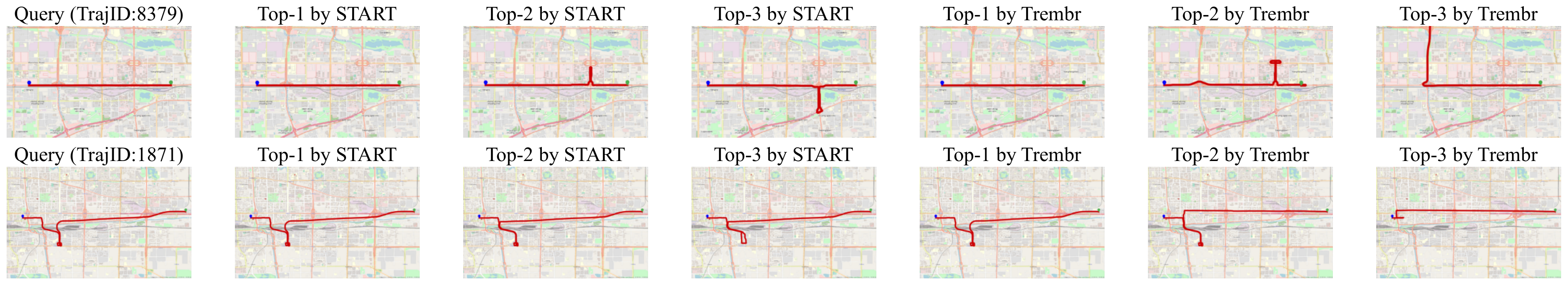}
    \vspace{-.3cm}
    \caption{\changed{Comparision of Top-3 Similar Trajectories Retrieved by \name and Trembr. (Map data $\copyright$ OpenStreetMap contributors, CC BY-SA.)}}
    \label{fig:top3}
    \vspace{-.6cm}
\label{exp:figknn}
\end{figure*}

In the experiments we set $N_{q}=10,000$, $N_{neg}=100,000$, select proportion $p_d=0.2$, time threshold $t_d=0.2$. Table~\ref{exp:tableresult} shows detailed results, and our model outperforms all baselines, especially in the mean rank (MR) metric. We attribute this to the fact that the representations learned by the model capture the travel semantics of the trajectories. In this way, the model can find the shape and semantically similar trajectories, an advantage that sequence-to-sequence models do not have. 


\textbf{\emph{(b) $k$-nearest Trajectory Search:}}
In the $k$-nearest search task, given a query trajectory, models need to find top-$k$ similar trajectories from the target database, ignoring the rank. Here, we use each query trajectory $\mathcal{T}_a$ in the query dataset $\mathcal{D_Q}$ to find the $k$-nearest-neighbors from the database $\mathcal{D_D}$ as ground truth. Then we construct the transformed detour dataset $\mathcal{D_Q}'$ from $\mathcal{D_Q}$ using the same method as above. For each transformed query $\mathcal{T}_a' \in \mathcal{D_Q}'$, we find the $k$-nearest-neighbors from database $\mathcal{D_D}$ and compare them to the ground truth. Since different selection proportions $p_d$ significantly change the generated trajectories, we vary $p_d$ from 0.1 to 0.5 to generate multi-data to evaluate the models. Figure~\ref{fig:proportions} shows the Precision of different models when the $p_d$ is varied and $k$ is fixed at 5. The Precision of all methods decreases as the selection proportion $p_d$ increases. \name always stays ahead and decreases more slowly, while Transformer, BERT, PIM-TF, and Toast perform less well. This is likely because the representations learned by the self-attention model are anisotropic~\cite{gao2019representation} and difficult to adapt to downstream tasks without fine-tuning. If we change the time threshold $t_d$, we obtain similar results not reported here.

\textbf{\emph{(c) Comparision of Top-3 Similar Trajectories:}}
To intuitively examine the search results of our proposed \name, we randomly select two trajectories and retrieve the top-3 similar trajectories using \name and Trembr, respectively, as shown in Figure~\ref{fig:top3}. The results show that \name can find diverse trajectories that are not exactly consistent with the query, but their overall trends (shape, OD, etc.) are similar. Compared with Trembr, the trajectory found by \name is closer to the query, while the result of Trembr deviates more from the query, especially the top-3 of query 8379. This illustrates the effectiveness of our proposed \name in capturing the global features and travel semantics of the trajectory.


\begin{figure}[t]
\centering
\includegraphics[width=1\columnwidth]{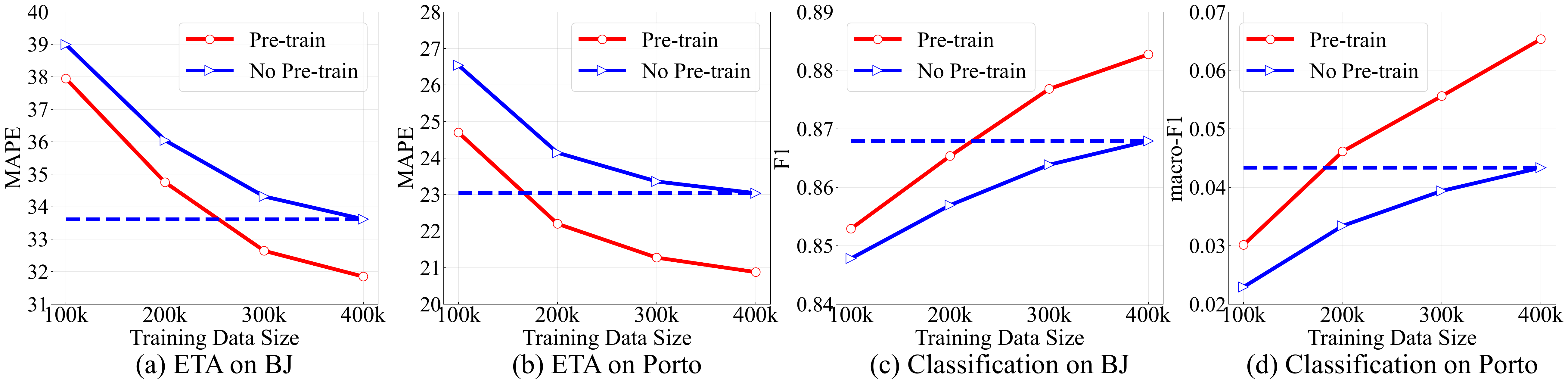}
    \vspace{-.6cm}
    \caption{Performance When Train Size Vary.}
    \label{fig:less}
    \vspace{-.5cm}
\label{exp:train_size}
\end{figure}

\subsection{Effect of Pre-training}
In this section, we verify the effectiveness of the two pre-training tasks we designed in two ways. One is to explore whether the training data size can be reduced by pre-training. The other is to investigate whether the pre-trained model can be transferred to other small datasets, even with a heterogeneous road network, to solve the problem of insufficient training data in many real-world applications.

\subsubsection{Performance Over Small Size Datasets}\label{exp:less}
One of the advantages of pre-training is that it can reduce the use of training data. We reduce the training data size for pre-training and fine-tuning and compare the proposed \name with the variant without pre-training (noted as \textit{No Pre-train}), \ie trained in a supervised manner. Figure~\ref{exp:train_size} shows performance on the entire test dataset of travel time estimation (ETA) and trajectory classification. We vary the size of the training data from 100k to 400k and train both \textit{No Pre-train} and \name. We find that the performance of both models improves with more labeled data, and \name consistently outperforms the \textit{No Pre-train} variant regardless of the training data size. Besides, as more data is used for pre-training, the performance of the model improves more significantly.  These experiments show that pre-training can effectively reduce the use of training data. 

\begin{table}[t]
  \centering
  \small
  \caption{Performance Comparison when transfer model across datasets.}
  \resizebox{\columnwidth}{!}{
    \begin{tabular}{c|ccc|ccc}
    \thickhline
          & \multicolumn{3}{c|}{Travel Time Estimation} & \multicolumn{3}{c}{Trajectory Classification} \\
    \hline
    Models & MAE   & MAPE(\%) & RMSE  & Micro-F1 & Macro-F1 & Recall@2 \\
    \hline
    \textit{No Pre-train \geolife} & 12.325 & 78.547 & 19.584 & 0.519 & 0.498 & 0.790 \\
    \hline
    \textit{Pre-train \geolife} & 11.980 & 73.489 & 18.613 & 0.568 & 0.571 & 0.814 \\
    \hline
    \textit{Porto-\name} & 10.455 & 65.371 & 18.024 & 0.623 & 0.619 & 0.832 \\
    \hline
    \textit{BJ-\name} & \textbf{9.995} & \textbf{64.331} & \textbf{17.183} & \textbf{0.669} & \textbf{0.665} & \textbf{0.887} \\
    \hline
    \textit{Porto-Trembr} & 15.200 &	80.294 & 23.223 & 0.507 & 0.468 & 0.728 \\
    \hline
    \textit{BJ-Trembr} & 14.851 & 79.239 &  23.109 & 0.512 & 0.486 & 0.741 \\
    \hline
    \thickhline
    \end{tabular}%
    }
  \label{tab:geolife}%
  \vspace{-.5cm}
\end{table}%

\subsubsection{Transfer Model Across Datasets}\label{exp:trans}
We transfer the model that pre-trained on a large dataset to another small dataset for fine-tuning, with the expectation that the knowledge learned from the large dataset will be transferred to the small one to solve the problem of insufficient training data. The small dataset is \geolife~\footnote{\url{https://research.microsoft.com/en-us/projects}}, a public dataset consisting of trajectories from 2007 to 2012 in Beijing. Since we need to perform \textit{map matching}, we only keep the trajectories with four transportation modes, including Car/Taxi, Walk, Bike, and Bus. In this way, we obtain 5,760 trajectories after data processing. In Table~\ref{tab:geolife}, we compare the performance of the following models: (1) \name trained directly or pre-trained with fine-tuned on \geolife (noted as \textit{No Pre-train \geolife}, \textit{Pre-train \geolife}), (2) \name pre-trained on \bj and \porto and fine-tuned on \geolife (noted as \textit{BJ-\name}, \textit{Porto-\name}), and (3) the best baseline Trembr pre-trained on \bj and \porto and fine-tuned on \geolife (noted as \textit{BJ-Trembr}, \textit{Porto-Trembr}). Since the source datasets are cab datasets, we use only the 882 Car/Taxi mode trajectories from the \geolife data for travel time prediction. The label for the trajectory classification is the four transportation modes.

We can conclude the following: (1) Direct pre-training on small datasets can also improve performance compared to non-pre-training. (2) Our proposed \name, whether pre-trained on \bj or \porto, outperforms the model pre-trained on \geolife, showing that pre-training can significantly improve performance on small datasets through knowledge transfer. The model pre-trained on \bj performs better than the model pre-trained on \porto because \bj and \geolife have the same road network. The parameters of our \gat layer are independent of the number of roads so that it can learn the road representations as long as the road network and the road features are given. Therefore, we can transfer \name to a heterogeneous road network dataset. In terms of performance improvement, we argue that the model learns the deep travel semantics of trajectories that are similar between different cities, thus enabling improvements. (3) When the pre-trained Trembr model is transferred to the \geolife dataset, performance is even worse. This confirms that the sequence-to-sequence model is unsuitable for transferring between datasets. Instead, our proposed \name is suitable for pre-training and transfer learning to solve the insufficient data problem.


\vspace{-.1cm}
\subsection{Ablation Study}\label{exp:ablation}
\begin{figure}[t]
\centering
\includegraphics[width=0.9\columnwidth]{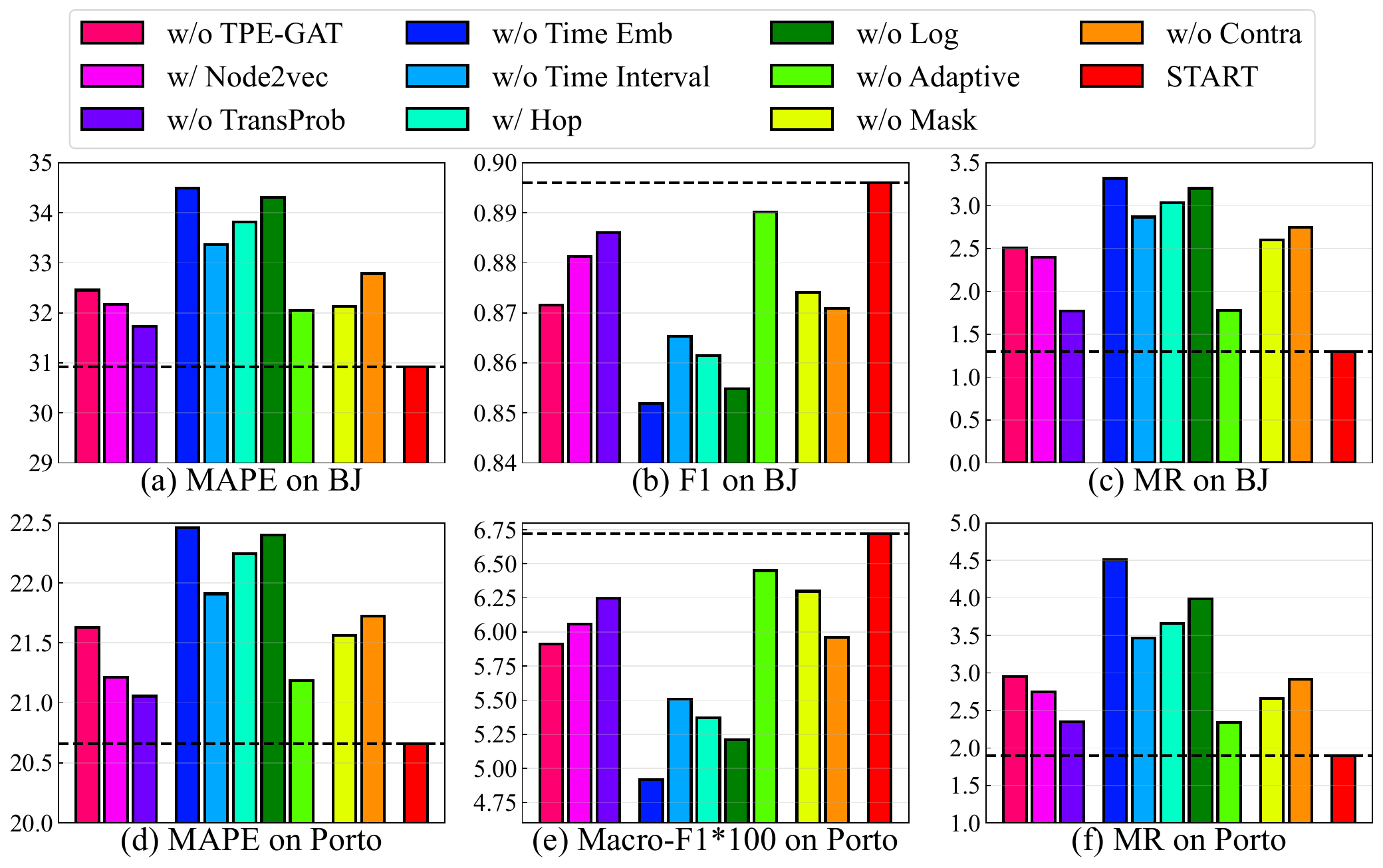}
    \vspace{-.3cm}    
    \caption{\changed{Ablation Study on \bj and \porto.}}
    \label{fig:abla}
    \vspace{-.6cm}
\label{exp:abla}
\end{figure}

To further investigate the effectiveness of each sub-module in \name, we conduct the following ablation experiments on both datasets. All experiments are repeated ten times and report the average results in Figure~\ref{fig:abla}. Due to space limitations, we show only one metric for each task. 


\subsubsection{Impact of Trajectory Pattern-Enhanced Graph Attention Layer} (a) \textit{w/o \gat}: this variant replaces the \gat with randomly initialized learnable road embeddings. (b) \textit{w/ Node2vec}: the variant replaces the \gat with learnable road embeddings initialized by node2vec~\cite{grover2016node2vec} algorithms. (c) \textit{w/o TransProb}: this variant removes the transfer probability matrix in the \gat. We can see that performance drops significantly without the \gat. The variant \textit{w/o TransProb} removes the transfer probability matrix so that the \gat degenerates to a standard GAT. This variant performs better than the variant \textit{w/ Node2vec} because compared with a standard GAT, the node2vec only focuses on the road network structure but ignores the road features. Moreover, this variant performs worse than the original model, reflecting that with transfer probabilities, the learned travel pattern enhanced-road representations are more valuable than the simple aggregation of the neighbors' features by a standard GAT.



\subsubsection{Impact of Time-Aware Trajectory Encoder Layer} (a) \textit{w/o Time Emb}: this variant drops the temporal embeddings ($\bm{t}_{mi}, \bm{t}_{di}$) to ignore the periodic temporal patterns. (b) \textit{w/o Time interval}: this variant drops the time interval matrix $\tilde{\bm{\Delta}}$. \changed{(c) \textit{w/ Hop}: this variant uses the number of hops between roads instead of the time interval to obtain the relative distance, \ie using $\delta _{i,j} = |i - j|$ instead of $\delta _{i,j} = |t_i - t_j|$. (d) \textit{w/o Log}: this variant replaces the logarithmic function that processes the time interval, \ie using $\delta_{i,j}' = {1}/{\delta_{i,j}}$ instead of $\delta_{i,j}' = {1}/{\log({\rm e} + \delta_{i,j})}$. (e) \textit{w/o Adaptive}: this variant drops the Eq.~\eqref{eq:learnable_timeinterval}, \ie $\tilde{\delta}_{i,j} = \delta_{i,j}' = {1}/{\log({\rm e} + \delta_{i,j})}$. In this way, the time interval matrix remains constant during the training process. We can see that the performance decreases significantly after neglecting the periodic temporal patterns (\ie \textit{w/o Time Emb}). It confirms the necessity of introducing periodic urban patterns. Besides, removing the time interval matrix leads to significant performance degradation. The variant \textit{w/ Hop} performs worse than \textit{w/o Time interval}, illustrating the importance of using the time interval between roads to measure the impacts among roads rather than using the hop distance. Similarly, the variant \textit{w/o Log} performs worse than \textit{w/o Time interval} because the inverse function changes too little at larger time intervals. The model performance also decreases if the matrix remains constant during the training process (\ie \textit{w/o Adaptive}), indicating the value of making the time interval matrix adaptive in model learning.}
    
\subsubsection{Impact of Self-Supervised Tasks} (a) \textit{w/o Mask}: this variant removes the span-masked loss and trains only with $\bm{\mathcal{L}}^{con}$. (b) \textit{w/o Contra}: this variant removes the contrastive loss and trains only with $\bm{\mathcal{L}}^{mask}$. Results demonstrate that both self-supervised pre-training tasks significantly affect the performance of downstream tasks.

\subsubsection{Impact of Data Augmentation Strategies} 
\begin{figure}[t]
\centering
\includegraphics[width=0.95\columnwidth]{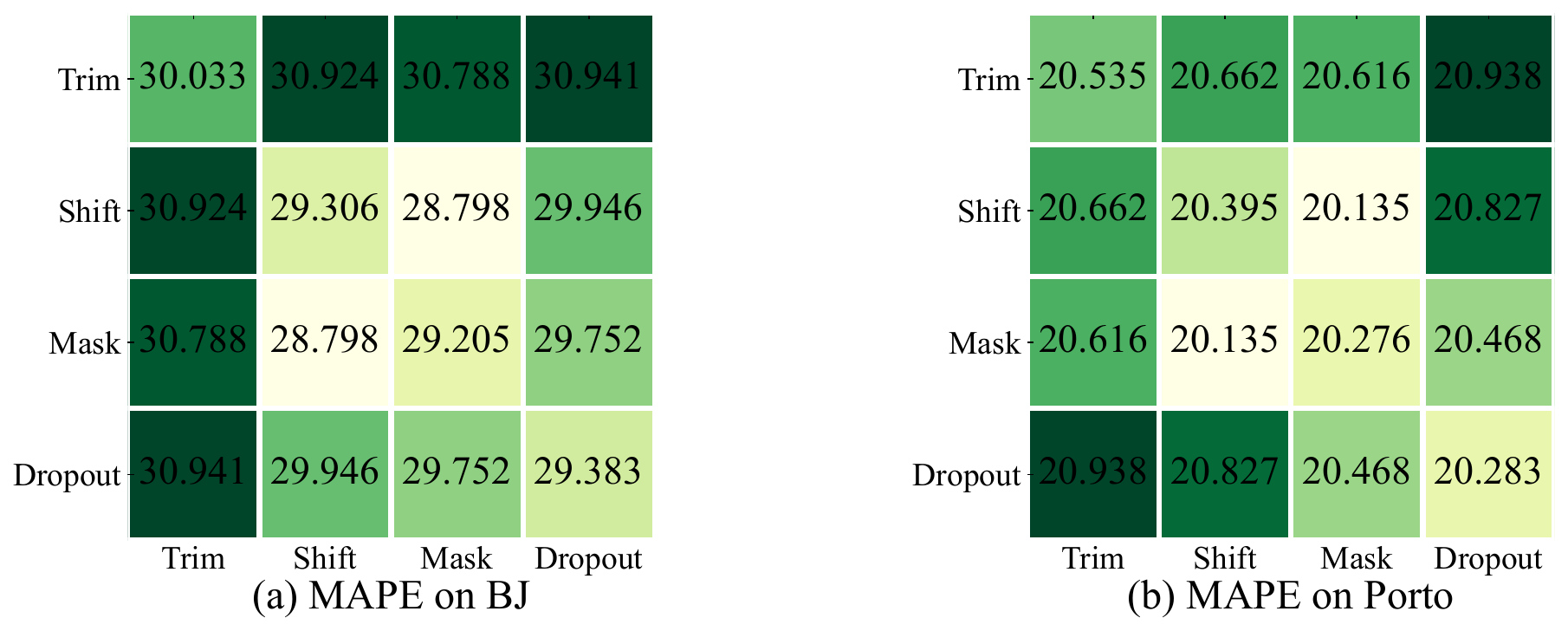}
    \vspace{-.2cm}
    \caption{MAPE for Different Data Augmentations.}
    \label{fig:data_augmentation}
    \vspace{-.3cm}
\end{figure}

Data augmentation strategies are central in contrastive learning to capture the spatial-temporal characteristics and travel semantics. We show performance with different pairs of methods to explore our proposed four trajectory data augmentation strategies. Due to space constraints, we only show the performance of travel time prediction. As shown in Figure~\ref{fig:data_augmentation}, we use a 4*4 grid to show the performance of different pairs of augmentation methods, where each row and column of the grid represents a data augmentation method. Note that the smaller the MAPE, \ie the lighter the color, the better the performance. We find that \textit{Temporal Shifting} and \textit{Road Segments Mask} perform best in this task. This shows that temporal regularities matter since both methods exhibit a change in the temporal dimension. Besides, \textit{Dropout} is a simple but efficient strategy that does not break the semantics of the trajectory.

\begin{figure}[t]
\centering
\includegraphics[width=0.95\columnwidth]{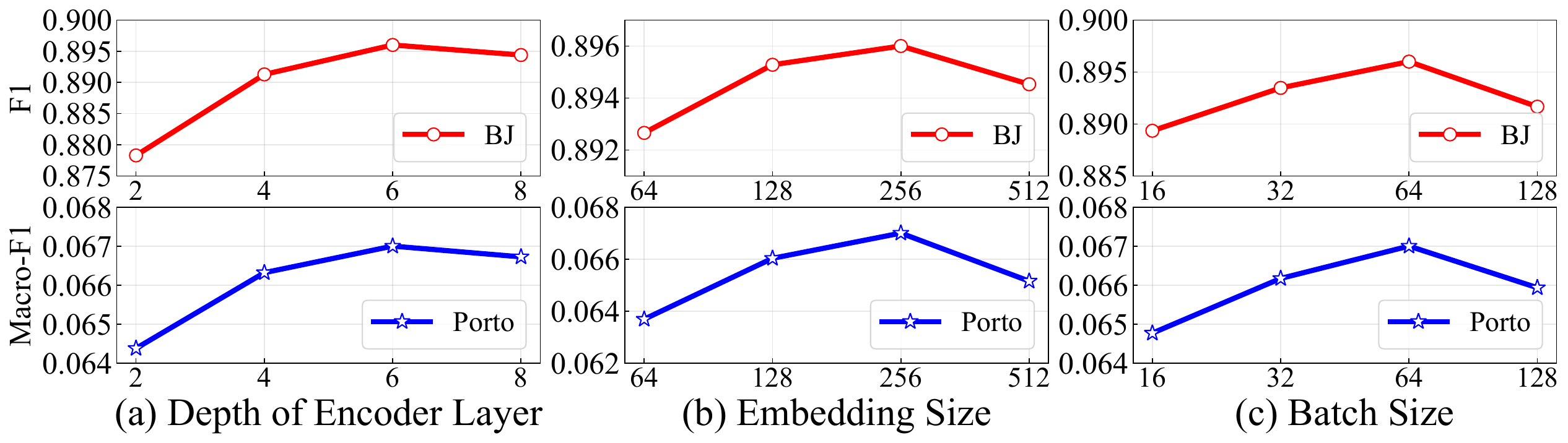}
    \vspace{-.2cm}
    \caption{Parameter Sensitivity Analysis.}
    \label{fig:para}
    \vspace{-.5cm}
\end{figure}


\subsection{Parameter Sensitivity}\label{exp:param}
We further conduct the parameter sensitivity analysis for critical hyperparameters, \eg encoder layers $L_2$, embedding size $d$, and batch size $N_b$ on both datasets. We report only the results of trajectory classification, and the results of the other tasks are similar. From Figure~\ref{fig:para}, we can see that the model performance initially improves with $d$ and $L_2$ increasing, but when they are too large, the performance deteriorates due to overfitting. Although previous studies have generally recommended larger batch sizes for contrastive learning~\cite{chen2020simple}, experiments have shown that model performance drops when batch sizes are too large. This may be due to a large batch introducing too many ``hard'' negative samples that differ minimally from the given anchor, \eg trajectories between the same ODs departing simultaneously. It is inappropriate to set these two semantically similar samples as a negative pair.


\subsection{Model Efficiency And Scalability}\label{exp:efficiency}

\begin{figure}[t]
\centering
\includegraphics[width=0.95\columnwidth]{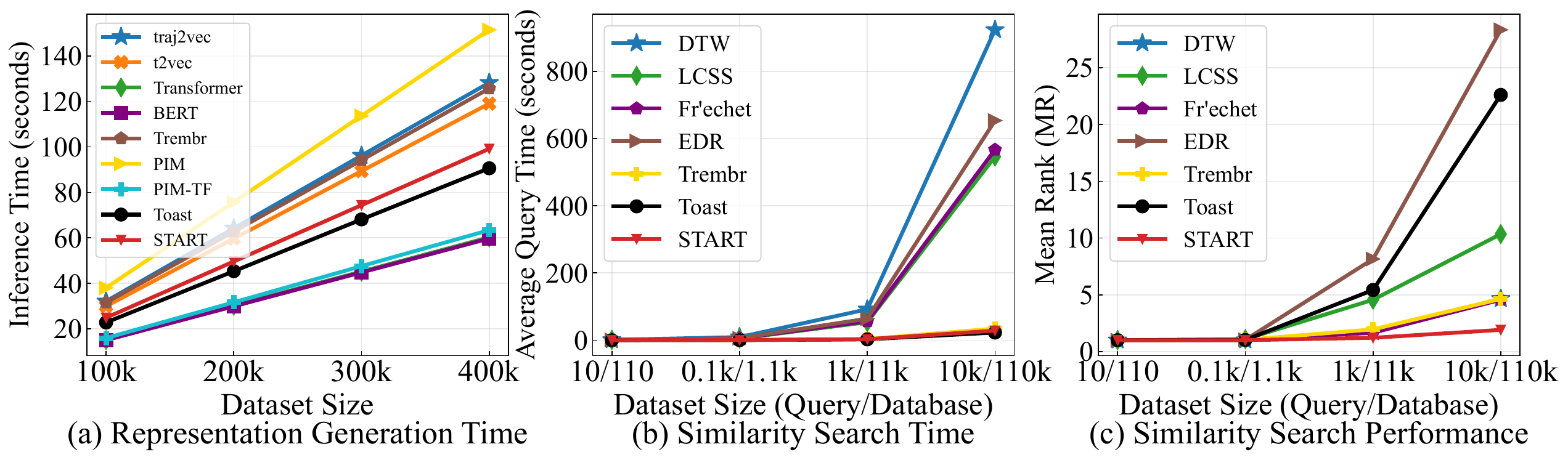}
    \vspace{-.2cm}
    \caption{Efficiency And Scalability Analysis.}
    \label{fig:efficiency}
    \vspace{-.4cm}
\end{figure}


Since \name is a pre-trained method for learning representations that can be trained offline, in practicality, we are more concerned with the time cost of encoding the trajectories into representations and applying them to downstream tasks. Here we use the default settings described in Section~\ref{exp:set}. We report only the results for \porto due to space limitations, and the results for \bj are similar.

First, we report the inference time of \name and baselines, \ie the time cost of embedding 100k-400k trajectories in Figure~\ref{fig:efficiency}(a). The results show that self-attention models are faster than RNN models. This is because RNN model need $O(L)$ sequential operations to process the trajectory while self-attention model only need $O(1)$. Here $L$ is the length of the trajectory. Besides, \name is slightly slower than other self-attention models because it introduces the \gat layer and the time interval matrix, a tradeoff between performance and efficiency. Even so, it takes only 25.8 seconds to encode 100,000 trajectories, and it can get even faster as the batch size gets larger during inference.

In addition, we also compare the time cost of similarity search. Figure~\ref{fig:efficiency}(b) shows the average time cost of a query for performing the most similar search with different query sizes and database sizes. The size of the query and detour dataset $N_q$ varies from 10 to 10,000, and the size of the negative samples $N_{neg}$ is ten times $N_q$. In addition to the two representative deep models Toast and Trembr, we also compare some traditional algorithms including Dynamic Time Warping (DTW)~\cite{yi1998efficient}, Longest Common SubSequence (LCSS)~\cite{vlachos2002discovering}, Fr'echet Distance~\cite{alt1995computing}, and Edit Distance on Real Sequence (EDR)~\cite{chen2005robust}. For the deep models, we sum the time cost of obtaining the representations and computing the similarities using the representations. 

We can see deep models are at least one order of magnitude faster than the traditional algorithms because the complexity of the traditional algorithms for computing the similarity is generally $O(L^2)$, while the deep models require only $O(d)$ complexity for computing the distance between the representations. Here $L$ is the length of the trajectory, and $d$ is the embedding size. Moreover, the deep models will be more efficient if we generate representations offline. The linear complexity makes \name scale well on large datasets. Moreover, as shown in Figure~\ref{fig:efficiency} (c), \name outperforms traditional algorithms on mean rank (MR) for search. This shows that \name is not only efficient but also can be used directly as a powerful metric for computing trajectory similarity without fine-tuning.

Finally, from the two experiments above, it appears that both the time for inference and the time for similarity search of \name increases linearly with the amount of data, which means \name can be scaled for large datasets.



\section{Related Work}

\subsection{Trajectory Representation Learning} \label{relate:trl}

\changed{Trajectory Representation Learning (TRL) is a powerful tool for spatial-temporal data analysis and management. TRL aims to convert raw trajectories into generic low-dimensional representation vectors that can be applied to various downstream tasks. Two of the earliest works that introduce the concept of trajectory representation learning into trajectory data management are t2vec~\cite{li2018deep} and traj2vec~\cite{yao2017trajectory}. T2vec~\cite{li2018deep} is trained by reconstructing high-sampling trajectories from low-sampling trajectories. Traj2vec~\cite{yao2017trajectory} transforms trajectories into feature sequences and trains a sequence-to-sequence (seq2seq) model based on reconstruction loss. Since then, many trajectory representation learning methods have been proposed for specific downstream tasks. For example, NEUTRAJ~\cite{yao2019computing}, Traj2SimVec~\cite{zhang2020trajectory}, and T3S~\cite{yang2021t3s} aim to learn trajectory representations for approximate trajectory similarity computation. In addition, DETECT~\cite{yue2019detect} and E2dtc~\cite{fang20212} build a seq2seq model trained with a reconstruction loss and a cluster-oriented loss to learn representations for trajectory clustering. GM-VSAE~\cite{abnormal} and D-TkDI~\cite{prank} learn trajectory representations for anomalous trajectory detection and path ranking, respectively.}



\changed{Most previous TRL works consider trajectories as sequences of locations, such as road segments, GPS sample points, or POI points while ignoring the corresponding temporal information. To the best of our knowledge, before our work, Trembr~\cite{fu2020trembr} was the only work that considered temporal information in self-supervised trajectory representation learning. Trembr is an RNN-based encoder-decoder model that considers the timestamps of each location in the decoding process. However, Trembr does not capture the periodic patterns of urban traffic or the irregular time intervals between trajectory samples. Our model explicitly incorporates the two temporal regularities into the trajectory representations, so it outperforms Trembr in the experiments.} In addition to GPS trajectories, some studies focus on other types of trajectories. TRED~\cite{zhou2020semi}, CTLTR~\cite{zhou2021contrastive}, and SelfTrip~\cite{gao2022self} are semi- or self-supervised representation methods for trip recommendations based on sparse POI check-in trajectories. 

\changed{In recent years, some two-stage methods have been proposed to learn generic trajectory representations for multiple downstream tasks~\cite{chen2021robust,yang2021unsupervised}. These methods first adopt a graph representation learning model to learn the road representation vectors and then use sequence learning models with self-supervised tasks to convert the road representation vectors in the same trajectory into the trajectory representations. For example, Toast~\cite{chen2021robust} and PIM~\cite{yang2021unsupervised} use node2vec~\cite{grover2016node2vec} to learn road representations and respectively use Transformer with masked prediction and RNN with mutual information maximization as self-supervised tasks to generate trajectory representations. Compared to our work, these two-stage methods consider trajectories as ordinary sequence data and thus ignore the temporal information. Besides, they only incorporate the static road network as spatial semantic information while ignoring the travel semantics, such as road visit frequencies.}

\subsection{Self-supervised Learning}
Self-supervised learning is a technique that enables learning with unlabeled data and has recently achieved remarkable success in various fields, such as computer vision~\cite{chen2020simple}, natural language processing~\cite{devlin2018bert, yan2021consert}, and data engineering~\cite{hgnnwn, wang2017community}. Self-supervised methods primarily include generative, predictive, and contrastive methods~\cite{yu2022self}. The generative methods learn representations based on reconstruction losses, such as some seq2seq models mentioned in Section~\ref{relate:trl}. The predictive methods construct labels based on the input data, such as BERT~\cite{devlin2018bert}, using the mask language prediction for training. The contrastive methods construct positive and negative samples and train the models to close the distance between positive pairs and push the distance between negative pairs. SimCLR~\cite{chen2020simple} is a contrastive learning method for visual representations using normalized temperature-scaled cross-entropy loss (NT-Xent) as training loss. SimCSE~\cite{gao2021simcse} uses standard dropout as noise to construct positive instances for sentence embeddings. ConSERT~\cite{yan2021consert} proposes four different types of contrastive learning data augmentation methods for learning sentence embeddings. Although several works have focused on the self-supervised learning of trajectories, our framework is the first to use both predictive and contrastive methods to capture the temporal regularities and travel semantics of the road-network constrained trajectory.

\section{Conclusion and Future work}

\changed{In this paper, we proposed a two-stage trajectory representation learning method, \name,  which incorporated temporal regularities and travel semantics into generic trajectory representation learning. Furthermore, we designed two self-supervised tasks to train our \name, which fully considered the spatial-temporal characteristics of trajectories. Extensive experiments on two large-scale datasets for three downstream tasks confirmed the superior performance of our proposed framework compared with the state-of-the-art baselines. The experiment results also demonstrated that our methods could be transferred across heterogeneous trajectory datasets, which was very useful for solving the problem of insufficient data.} 

\changed{In the future, we plan to explore more data augmentation techniques for contrastive learning according to the specific downstream tasks and extend the proposed framework to other categories of trajectory data, such as POI check-in trajectories, to support more applications.}

\section*{Acknowledgment}
This work was supported by the National Key R\&D Program of China (2019YFB2103203). Prof. Wang’s work was supported by the National Natural Science Foundation of China (No. 82161148011, 72222022, 72171013), and the DiDi Gaia Collaborative Research Funds.

\clearpage

\bibliographystyle{IEEEtran}
\bibliography{references}


\end{document}